\pdfoutput=1

\documentclass[11pt]{article}

\usepackage[final]{acl}
\usepackage{acl}

\usepackage{times}
\usepackage{latexsym}

\usepackage[T1]{fontenc}

\usepackage[utf8]{inputenc}

\usepackage{microtype}

\usepackage{inconsolata}

\usepackage{graphicx}
\usepackage{subcaption}
\usepackage{overpic}
\usepackage{tikz}

\usepackage{amssymb}
\usepackage{amsfonts}
\usepackage{mathtools}
\usepackage[capitalize, noabbrev]{cleveref}
\usepackage{amsmath}
\usepackage{diagbox}
\DeclareMathOperator*{\argmin}{arg\,min}
\usepackage{algorithm}       
\usepackage{algpseudocode}
\usepackage{float}

\usepackage{booktabs}   
\usepackage{colortbl}   
\usepackage{array}      
\usepackage{lscape}     
\usepackage{adjustbox}  
\usepackage{threeparttable} 
\usepackage{makecell}
\usepackage{amsmath}
\usepackage{multirow}

\usepackage[most]{tcolorbox}
\usepackage{mdframed}

\algrenewcommand\algorithmicrequire{\textbf{Input:}}
\algrenewcommand\algorithmicensure{\textbf{Output:}}

\newcommand{\red}[1]{{\color{red}#1}}
\newcommand{\blue}[1]{{\color{blue}#1}}

\definecolor{box}{RGB}{255, 239, 239}

\newcommand{\bH}{\mathbf{H}}
\newcommand{\bL}{\mathbf{L}}
\newcommand{\bQ}{\mathbf{Q}}
\newcommand{\bR}{\mathbf{R}}
\newcommand{\bS}{\mathbf{S}}
\newcommand{\bX}{\mathbf{X}}
\newcommand{\bW}{\mathbf{W}}

\newcommand{\field}[1]{\mathbb{#1}}
\newcommand{\fR}{\field{R}}

\title{Assigning Distinct Roles to Quantized and Low-Rank Matrices \\ Toward Optimal Weight Decomposition}

\author{
  Yoonjun Cho$^{1}$ \;
  Soeun Kim$^{2}$ \;
  Dongjae Jeon$^{1}$ \;
  Kyelim Lee$^{2}$ \;
  Beomsoo Lee$^{3}$ \;
  Albert No$^{2}$\thanks{
  Correspondence to: Albert No <albertno@yonsei.ac.kr>}\\[0.5em]  
  \normalsize{$^{1}$Department of Computer Science, Yonsei University} \\
  \normalsize{$^{2}$Department of Artificial Intelligence, Yonsei University} \\
  \normalsize{$^{3}$Department of Electronic \& Electrical Convergence Engineering, Hongik University}
}

\begin{document}
\maketitle
\setcounter{footnote}{0}
\renewcommand{\thefootnote}{\arabic{footnote}}

\begin{abstract}
Decomposing weight matrices into quantization and low-rank components ($\mathbf{W} \approx \mathbf{Q} + \mathbf{L}\mathbf{R}$)
is a widely used technique for compressing large language models (LLMs). 
Existing joint optimization methods iteratively alternate between quantization and low-rank approximation. 
However, these methods tend to prioritize one component at the expense of the other, resulting in suboptimal decompositions that fail to leverage each component's unique strengths.
In this work, we introduce Outlier-Driven Low-Rank Initialization (ODLRI), which assigns low-rank components the specific role of capturing activation-sensitive weights.
This structured decomposition mitigates outliers' negative impact on quantization,
enabling more effective balance between quantization and low-rank approximation. 
Experiments on Llama2 (7B, 13B, 70B), Llama3-8B, and Mistral-7B demonstrate that incorporating ODLRI into the joint optimization framework consistently reduces activation-aware error, minimizes quantization scale, and improves perplexity and zero-shot accuracy in low-bit settings.
\end{abstract}

\section{Introduction}
Quantization~\citep{polino2018model, jacob2018quantization, nagel2021white} and weight matrix factorization~\citep{GOLUB1987317, saha2023matrix} 
are two widely used techniques for compressing large language models (LLMs) 
to enable efficient inference on resource-constrained hardware.
Post-training quantization (PTQ) reduces model size and computational cost by mapping high-precision weights to lower-bit representations~\citep{tao2022compression, bai2022towards, dettmers2022gpt3, shaoomniquant},
while matrix factorization approximates weight matrices with compact factored representations~\citep{li2023losparse, gao-etal-2024-adaptive}.
Recent joint optimization approaches combine these methods to achieve extreme compression,
representing weights as the sum of a quantization matrix and a low-rank component: $\bW\approx \bQ+\bL\bR$~\citep{saha2024compressing, li2024loftq, guolq}.

Joint optimization approaches minimize representation error by iteratively alternating between quantization and low-rank approximation.
These methods typically adopt either a quantize-first strategy~\citep{saha2024compressing, li2024loftq} or a low-rank-first strategy~\citep{guolq}.
While these differ in iteration ordering, they can be equivalently understood as distinct initialization choices for the low-rank components:
the quantize-first approach initializes $\bL\bR$ to zero, while the low-rank-first approach initializes them using factorized weights.

Critically, existing methods treat these different orderings merely as preparatory steps, assuming iterative updates will naturally converge to optimal solutions.
However, viewing this process through the lens of initialization reveals an underexplored aspect of how initialization strategies might fundamentally affect weight decomposition quality.

In this work, we investigate the role of initialization in joint optimization.
Our analysis shows that different initializations lead to distinct solution spaces, with components maintaining persistent roles throughout optimization.
Quantize-first methods treat low-rank components as error correction terms, while low-rank-first methods preserve them as the primary weight representation.
This finding highlights that initialization fundamentally determines the role assignment between quantization and low-rank components, raising a key question:
\begin{center}
\begin{tcolorbox}[colframe=black, colback=pink!20, boxrule=1pt, width=\linewidth, arc=5pt]
{\it  What is the optimal initialization strategy for decomposing weights into quantized and low-rank matrices?
}
\end{tcolorbox}
\end{center}

Recent works have shown that quantization errors are pronounced for weights associated with activation outliers, as extreme activations amplify weight sensitivity~\citep{dettmers2023spqr, lin2024awq, lee2024owq, huang2024slim, kimsqueezellm}.
Building on this insight, we introduce Outlier-Driven Low-Rank Initialization (ODLRI), which assigns a specific role to the low-rank component to capture these salient weights while using the quantized matrix to express the residuals.
By handling outlier-sensitive weights through the low-rank component, our approach stabilizes quantization and enables more precise weight decomposition.

Through extensive experiments on Llama2 (7B, 13B, 70B)~\citep{touvron2023llama}, Llama3-8B~\citep{dubey2024llama}, and Mistral-7B~\citep{jiang2023mistral},
we demonstrate that incorporating ODLRI into the joint optimization framework consistently improves perplexity and zero-shot accuracy across extreme low-bit settings.
Our analysis shows that ODLRI reduces activation-aware error, minimizes quantization scale, and enhances model performance,
highlighting the importance of structured initialization in low-bit quantization.
These results present a principled approach for stable and efficient LLM compression, and provide new insights into the effective decomposition of quantization and low-rank matrices.

We summarize our contributions as follows:
\begin{itemize}
\vspace{-0.1em}
\item We propose a unified framework for expressing iterative joint optimization algorithms by introducing the concept of initialization of low-rank components.
\vspace{-0.1em}
\item We analyze the impact of initialization on iterative joint optimization algorithms, revealing the suboptimality of conventional approaches. 
\item We propose Outlier-Driven Low-Rank Initialization (ODLRI), which assigns a specific role to the low-rank component $\bL\bR$ to capture salient weights while using the quantization matrix $\bQ$ for the remaining weights.
\vspace{-0.1em}
\item Comprehensive experiments demonstrate that incorporating ODLRI reduces activation-aware error, minimizes quantization scale, and improves perplexity and zero-shot accuracy.

\end{itemize}

\section{Preliminaries}



\subsection{Post-Training Quantization}

Early post training quantization (PTQ) methods~\citep{jacob2018quantization, nagel2021white, dettmers2022gpt3, shaoomniquant} mainly focus 
on minimizing the direct quantization error for a given weight matrix $\bW\in\fR^{m\times n}$, optimizing:
\begin{align*} 
    \argmin_{\bQ} \|\bW-\bQ\|^2_\text{F}, 
\end{align*}
where $\bQ$ is obtained by rounding weights to the nearest discrete values~\citep{banner2019post, Stock2020And, wu2020integer}.
However, this naive rounding approach often results in significant accuracy degradation. This problem is particularly pronounced in large-scale LLMs, where even small perturbations in weights can propagate across layers, gradually accumulating and worsening errors.

To mitigate this issue, activation-aware PTQ methods~\citep{9009512, nagel2020up, librecq, li2024norm} incorporate a calibration dataset $\bX\in\fR^{n\times d}$
to account for the interaction between weight quantization and activations:
\begin{align*} 
    \argmin_{\bQ} \|(\bW-\bQ)\bX\|^2_\text{F}.
\end{align*}

This formulation ensures that weight quantization preserves the statistical behavior of activations, and leads to improved model performance.
OPTQ~\citep{frantar2023optq} refines activation-aware PTQ by introducing error feedback, which reduces cumulative quantization errors for effective low-bit quantization.
QuIP~\citep{chee2023quip} and QuIP\#~\citep{tsengquip} further improve robustness through incoherence processing, applying orthogonal transformations to $\bW$
and its local Hessian to reduce correlations, making 2-bit quantization more effective.

In addition, recent studies show that effectively controlling a small portion of activation outlier-sensitive weights can significantly enhance quantization performance.
SpQR~\citep{dettmers2023spqr} retains these critical salient weights in higher precision, while AWQ employs per-channel scaling to protect them, effectively mitigating the adverse effects of activation outliers during quantization.

\subsection{Weight Matrix Factorization}
Matrix factorization decomposes a matrix into low-rank components for efficient representation and computation~\citep{GOLUB1987317, saha2023matrix}.
Recently, it has been applied to LLM compression by approximating weight matrices with low-rank representations~\citep{li2023losparse, gao-etal-2024-adaptive}.
Given a weight matrix $\mathbf{W} \in \mathbb{R}^{m \times n}$,
these methods decompose it into lower-dimensional matrices, $\mathbf{L} \in \mathbb{R}^{m \times r}$ and $\mathbf{R} \in \mathbb{R}^{r \times n}$, by minimizing:
\begin{align*}
\argmin_{\bL, \bR} \|\bW-\bL\bR\|^2_\text{F}.
\end{align*}

By reducing the rank $r$, low-rank decomposition, such as Singular Value Decomposition (SVD) effectively reduces storage and computational costs while preserving key weight structures.

Activation-aware factorization further incorporates activation statistics to refine weight representation, keeping the decomposition aligned with the model’s actual computational behavior.
ASVD~\citep{yuan2023asvd} introduces a diagonal scaling matrix to normalize activations, improving numerical stability,
while SVD-LLM~\citep{wang2025svdllm} applies truncation-aware data whitening via Cholesky decomposition.

\subsection{Quantization Error Reconstruction}\label{subsec:qer}
Empirical studies suggest that quantization errors $\bW-\bQ$ often exhibit a low-rank structure~\citep{yao2024exploring}.
This observation has led to hybrid approaches that first quantize the weight matrix and then approximate the resulting quantization error using a low-rank term, which serves as an error compensation mechanism:
\begin{align*}
\bQ =& \argmin_{\bQ} \|\bW-\bQ\|^2_\text{F}\\
\bL, \bR =& \argmin_{\bL, \bR} \|\bW-\bQ-\bL\bR\|^2_\text{F}.
\end{align*}
Here, $\bQ$ represents the quantized weight matrix, forming the major representation of $\bW$,
while $\bL\bR$ provide a low-rank approximation of the residual quantization error~\citep{liu2024eora}.

ZeroQuant-V2~\citep{yao2024exploring} estimates $\bL$ and $\bR$ by applying SVD to the residual error $\bW-\bQ$.
Subsequent methods further refine this process by incorporating activation-aware adjustments:
LQER~\citep{zhanglqer} introduces a diagonal scaling matrix derived from activations,
while QERA~\citep{zhang2025qera} improves error reconstruction by leveraging an input-space autocorrelation matrix, and provides theoretical guarantees for enhanced performance.

Unlike Q-LoRA~\citep{dettmers2023qlora} and other quantization-based parameter-efficient fine-tuning (Q-PEFT) methods that focus on weight representations optimized for fine-tuning, quantization error reconstruction methods solely aim for efficient weight representation.

\subsection{Jointly Optimized Quantization and Low Rank Approximation}
A natural extension of quantization error reconstruction is to jointly optimize both the quantized component 
$\bQ$ and the low-rank component $\bL\bR$, leading to the following formulation:
\begin{align*}
\bQ, \bL, \bR= \argmin_{\bQ, \bL, \bR} \|(\bW-\bQ-\bL\bR)\|^2_\text{F}.
\end{align*}

Unlike the two-stage approach that applies quantization followed by low-rank error correction once, joint optimization methods often iteratively alternate between quantization and low-rank approximation, which generally leads to better performance.
These methods typically adopt either a quantize-first strategy~\citep{saha2024compressing, li2024loftq} or a low-rank-first strategy~\citep{guolq}, distinguished by their iteration ordering.

Rather than distinguishing them by iteration sequences, we propose interpreting these approaches through their initialization of low-rank components. 
Specifically, the quantize-first approach initializes $\bL\bR$ to zero, whereas the low-rank-first approach initializes them to factorized weights. 
Our perspective enables us to reframe joint optimization methods within a unified view in~\cref{alg:qer1}.

\begin{algorithm}[h]
\caption{Joint $\bQ + \bL\bR$ Optimization}
\label{alg:qer1}
\begin{algorithmic}[1]
\Require Pretrained weight $\mathbf{W}$, Num of iterations $T$
\Ensure $\mathbf{Q}_T , \mathbf{L}_T , \mathbf{R}_T$

\State $\mathbf{L}_0, \mathbf{R}_0 \gets \textcolor{Orange}{\textbf{Initialize}}$
\For {$t = 1$ to $T$}
    \State $\mathbf{Q}_t \gets \textcolor{blue}{\textbf{Quantize}}(\mathbf{W} - \mathbf{L}_{t-1} \mathbf{R}_{t-1})$
    \State $\mathbf{L}_t, \mathbf{R}_t \gets \textcolor{red}{\textbf{LRApprox}}(\mathbf{W} - \mathbf{Q}_t)$
\EndFor
\Statex {\bf Return:} $\widehat\bW = \mathbf{Q}_T + \mathbf{L}_T\cdot\mathbf{R}_T$
\end{algorithmic}
\end{algorithm}

Through~\cref{alg:qer1}, we gain deeper insights into the joint optimization procedure.
This framework not only allows us to explore algorithmic variations by examining different initialization strategies but also suggests the potential role of initialization choices in weight decomposition.

CALDERA~\citep{saha2024compressing}, a weight-only PTQ method jointly optimizes $\bQ$, $\bL$, and $\bR$ in an activation-aware manner, has shown significant improvements in quantization quality, particularly in extreme low-bit settings. 
Despite its outstanding performance, CALDERA solely examines its approach by initializing low-rank components to zero, neglecting the potential impact of initialization on performance. 
Within this CALDERA framework, we further concentrate on how different initialization strategies for the low-rank component affect performance, aiming to shed light on its role in the overall optimization process.

\begin{figure*}[!t]
    \centering
    \begin{minipage}{1\textwidth}
        \centering
        \includegraphics[width=\textwidth]{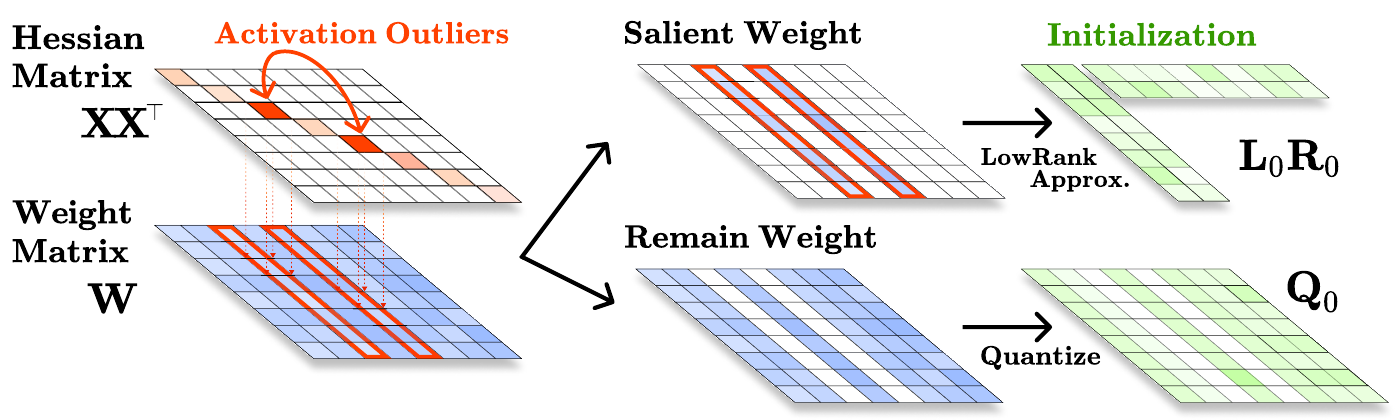} 
        \caption{\textbf{Outlier-Driven Low-Rank Initialization (ODLRI) Framework.} 
        ODLRI decomposes the weight matrix $\bW$ by first identifying salient weights, corresponding to activation outliers, using the diagonal of the Hessian.
        These salient weights are then approximated via low-rank decomposition, 
         producing $\mathbf{L}_0$ and $\mathbf{R}_0$, while the remaining weights are quantized. 
         This decomposition serves as initialization for the iterative joint optimization of $\mathbf{Q}+\mathbf{L}\mathbf{R}$.
        }
        \label{fig:method}
    \end{minipage}
    \hfill 
\end{figure*}
\section{Rethinking Joint Q+LR Optimization}
\subsection{Dependence on LR Initialization}

To examine the role of initialization in joint optimization, we evaluate CALDERA~\citep{saha2024compressing} under two distinct initialization strategies for $\bL\bR$: zero initialization and matrix factorization-based initialization.
For analyzing the contribution of quantized representation and low-rank components to $\mathbf{W}\mathbf{X}$, we measure $\|\mathbf{Q}\mathbf{X}\|$ and $\|\mathbf{L}\mathbf{R}\mathbf{X}\|$ at both the initial and final stage of iteration.

Surprisingly, as shown in~\Cref{tab:lrinit_order_norm}, when $\bL$, $\bR$ are initialized to zero, we observe that $\bQ$ persistently reconstructs $\bW$,
while $\bL\bR$ serves as a residual correction, closely resembling quantization error reconstruction.
Conversely, initializing $\bL, \bR$ with a matrix factorization of $\bW$ leads to a reversed role assignment, where $\bL\bR$ captures most of $\bW$, and $\bQ$ quantizes the residuals throughout the iteration.

\renewcommand{\arraystretch}{1.2}  
\setlength{\tabcolsep}{5pt}  
\begin{table}[!ht]
    \centering
    \Large
    \resizebox{\columnwidth}{!}{  
    \begin{tabular}{c | cc | cc}
        \toprule
        \multirow{1}{*}{\makecell{$\bL\bR$ Initialization}}
        & \multicolumn{2}{c|}{\textbf{0}}  
        & \multicolumn{2}{c}{LRApprox($\mathbf{W}$)} \\  
        \midrule
        \diagbox{}{}
        &$\|\bQ\bX\|$ & $\|\bL\bR\bX\|$
        & $\|\bQ\bX\|$ & $\|\bL\bR\bX\|$ \\
        \midrule
        First Iteration  & 0.999  & 0.014  & 0.158 & 0.915 \\
        Last Iteration & 0.961  & 0.073  & 0.401 & 0.664 \\
        \bottomrule
    \end{tabular}
    } 
    \caption{
    {Effect of $\bL\bR$ Initialization in CALDERA.}
    Activation norms $\|\bQ\bX\|$ and $\|\bL\bR\bX\|$  are reported at the first and last iterations for the Layer 1 Key Projection matrix of Llama2-7B over 15 iterations.
    Norms are normalized by $\|\bW\bX\|$ (i.e., $\|\bL\bR\bX\| / \|\bW\bX\|$). More results are provided in~\Cref{app: Effect of LR initialization}.}
    \label{tab:lrinit_order_norm}
\end{table}

This finding reveals that joint optimization outcomes are highly sensitive to initialization choices,
which ultimately determine whether quantization or matrix factorization dominates the final representation. 
While existing methods have defaulted to either zero initialization or low-rank approximation of $\mathbf{W}$ (i.e., LRApprox($\mathbf{W}$)), the optimality of these conventional approaches remains unexplored. 
Rather than being restricted to these established choices, we ask:

\begin{center}
\begin{tcolorbox}[colframe=black, colback=pink!20, boxrule=1pt, width=\linewidth, arc=5pt]
\textit{
How should we initialize $\mathbf{L}, \mathbf{R}$ to achieve optimal decomposition of $\mathbf{W}$ into $\mathbf{Q}+\mathbf{L}\mathbf{R}$?
}
\end{tcolorbox}
\end{center}

\subsection{Outlier-Driven Low-Rank Initialization}

For optimal decomposition of $\bW\approx \bQ+\bL\bR$, we assign distinct roles to quantization and low-rank approximation based on their properties.
Quantization is highly sensitive to activation outliers, as extreme activations amplify weight sensitivity,
leading to discretization errors that degrade model performance~\citep{lin2024awq, dettmers2023spqr}. In contrast, the low-rank component utilizes a product formulation of two low-bit factors, effectively yielding a higher bit representation than quantization.
To leverage the enhanced representational capacity of the low-rank component, we structure $\bL\bR$ to explicitly capture salient weights before quantization, rather than treating it as a post-hoc correction term.
This results in salient weights being absorbed into the low-rank component, allowing $\bQ$ to operate on a smoother, more uniform residual, ultimately improving quantization efficiency.

To explicitly assign the low-rank component to capturing salient weights, we introduce \textit{Outlier-Driven Low-Rank Initialization (ODLRI)}, as illustrated in~\Cref{fig:method}.
More precisely, we first decompose the activation matrix $\bX$ into two components:
\begin{align*}
    \bX = \bX_{\text{o}} + \bX_{\text{r}},
\end{align*}
where $\bX_{\text{o}}\in\fR^{n\times d}$ contains only the top-$k$ activation channels (outliers) with the highest norms,
the remaining channels are set to zero. 
These top-$k$ activation channels are identified by analyzing the diagonal entries of the Hessian, computed as $\bH = \bX\bX^\top$.
The remaining activations (non-outlier components) are then captured by $\bX_{\text{r}}$.

Although setting $k=r$ (where $\mathbf{L} \in \mathbb{R}^{m \times r}$ and $\mathbf{R} \in \mathbb{R}^{r \times n}$) would maximize the use of low-rank approximation,
we intentionally choose $k<r$ to focus aggressively on outlier-related structures rather than broadly approximating the entire weight distribution.
This targeted selection ensures that the low-rank component prioritizes the most activation-sensitive elements,
refining the decomposition and enhancing quantization robustness.

We then initialize $\bL$ and $\bR$ via outlier-aware matrix factorization, focusing on high-variance activation directions that are likely to induce quantization errors. Specifically, we solve the following truncated optimization problem:
\begin{align*}
\bL_0, \bR_0 =\argmin_{\bL, \bR} \|(\bW-\bL\bR)\bH_{\text{o}}(\bW-\bL\bR)^\top\|,
\end{align*}
where $\bH_{\text{o}} = \bX_{\text{o}}\bX_{\text{o}}^\top$ captures the covariance of outlier-sensitive channels.
This objective ensures that $\bL_0\bR_0$ prioritizes reconstructing weight directions that interact strongly with outlier channels in the activation distribution.
A detailed algorithm for solving this optimization is provided in~\Cref{app: Outlier sensitive Weight SVD for LR Initialization},
and the specific selection of $k$ is described in \Cref{app: Rank-Dependent Outlier Selection}.

\begin{figure*}[!htb]
    \centering
    \begin{subfigure}{0.32\textwidth}
        \centering
        \includegraphics[width=\linewidth]{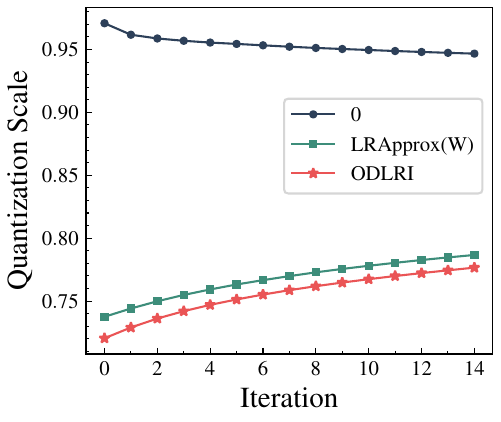}
        \caption{Key projection}
        \label{fig:q_scales_10_Key}
    \end{subfigure}
    \begin{subfigure}{0.32\textwidth}
        \centering
        \includegraphics[width=\linewidth]{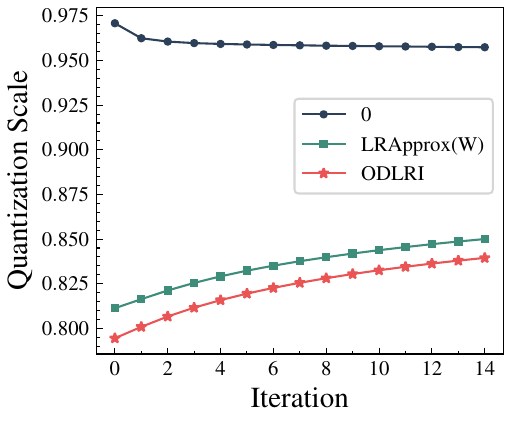}
        \caption{Value projection}
        \label{fig:q_scales_10_Value}
    \end{subfigure}
    \begin{subfigure}{0.32\textwidth}
        \centering
        \includegraphics[width=\linewidth]{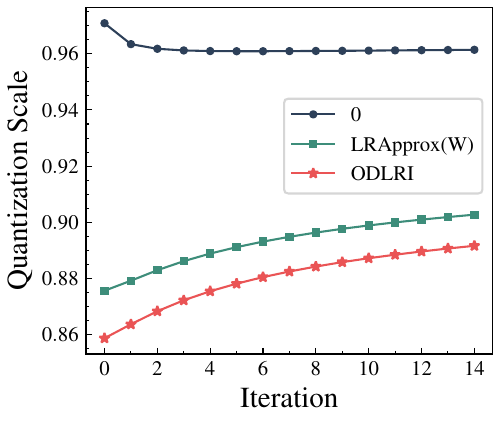}
        \caption{Down Projection}
        \label{fig:q_scales_10_Down}
    \end{subfigure}

    \caption{\textbf{Quantization Scale across different initialization strategies}.
   We present the quantization scale over 15 iterations, where both $\bL$ and $\bR$ are quantized to 4-Bit at rank 256.
    The three subplots display results for the Key (left), Value (middle), and Down (right) projection layers in Layer 10 of Llama2-7B.
    ODLRI (red stars) consistently achieves the lowest quantization scale, highlighting its effectiveness in low-bit quantization.
    Additional results are provided in the~\Cref{app: scale and error}}
    \label{fig:qscale_comp}
\end{figure*}
\begin{figure*}[!htb]
    \centering
    \begin{subfigure}{0.32\textwidth}
        \centering
        \includegraphics[width=\linewidth]{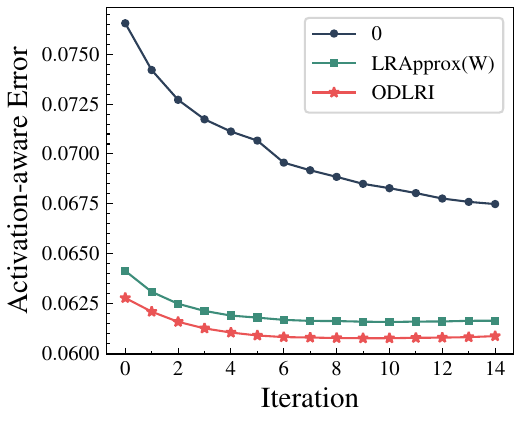}
        \caption{Key projection}
        \label{fig:error_comp_10_key}
    \end{subfigure}
    \begin{subfigure}{0.32\textwidth}
        \centering
        \includegraphics[width=\linewidth]{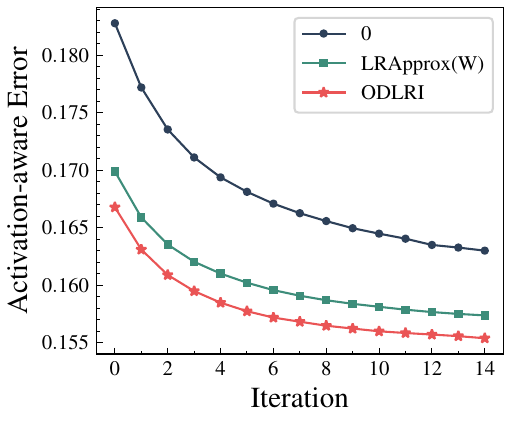}
        \caption{Value projection}
        \label{fig:error_comp_10_value}
    \end{subfigure}
    \begin{subfigure}{0.32\textwidth}
        \centering
        \includegraphics[width=\linewidth]{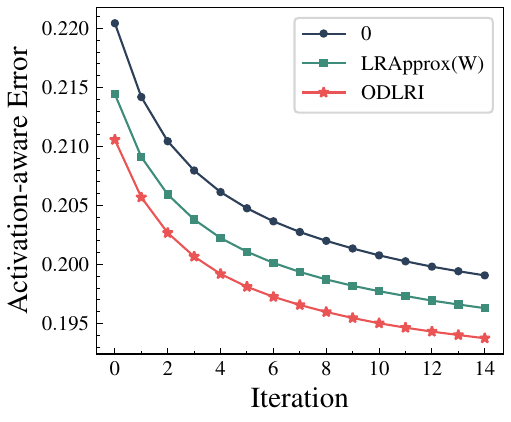}
        \caption{Down Projection}
        \label{fig:error_comp_10_down}
    \end{subfigure}

    \caption{\textbf{Activation-aware Error across different initialization strategies}.
    We present the normalized activation-aware error ${\|(\bW - \bQ - \bL \bR)\bX\|^2_\text{F}} / {\|\bW\bX\|^2_\text{F}}$
    over 15 iterations, where both $\bL$ and $\bR$ are quantized to 4-Bit at rank 256.
    The three subplots display results for the Key (left), Value (middle), and Down (right) projection layers in Layer 10 of Llama2-7B.
    ODLRI (red stars) consistently achieves the lowest error, demonstrating its effectiveness in accurately representing weights.
    Additional results are provided in the~\Cref{app: scale and error}}
    \label{fig:error_comp}
\end{figure*}

\section{Experiment}

\subsection{Experimental Setup} \label{subsec:exp_setup}
To evaluate the effectiveness of Outlier-Driven Low-Rank Initialization (ODLRI), we integrate it into CALDERA~\citep{saha2024compressing},
an iterative joint optimization framework for extreme low-bit quantization of $\bQ, \bL, \bR$.
By default, CALDERA first quantizes the weight matrix and then optimizes a low-rank component, effectively treating $\bL, \bR$ as zero-initialized correction factors.

In contrast, ODLRI replaces this zero-initialization with an outlier-aware initialization,
ensuring that salient weights are explicitly modeled in the low-rank component before quantization.
We conduct a comprehensive comparison against CALDERA's standard setup to quantify the impact of $\bL\bR$ initialization on final accuracy and stability.

\paragraph{Quantization Setup.}
We integrate ODLRI into CALDERA while keeping the quantization configuration largely unchanged.
For \textbf{\blue{Quantize}}, we use a 2-Bit quantization scheme implemented via QuIP\#~\citep{tsengquip}, which employs an E8 lattice codebook for stable 2-Bit quantization.
For \textbf{\red{LRApprox}}, we experiment with both 16-Bit and 4-Bit precision, where the 16-Bit setting remains unquantized, while the 4-Bit setting undergoes quantization. 
For 4-Bit precision, we adjust $\bL$ and $\bR$ via the LPLR iterative algorithm~\citep{saha2023matrix}.

CALDERA performs outer iterations alternating between \textbf{\blue{Quantize}} and \textbf{\red{LRApprox}},
along with inner iterations for the LPLR procedure.
We follow CALDERA's default configuration, running 15 outer iterations, and 10 inner iterations when the $\bL, \bR$ components are quantized to 4-Bit.

\renewcommand{\arraystretch}{0.9}  
\setlength{\tabcolsep}{4pt}  
\definecolor{Gray}{gray}{0.9}
\begin{table*}[ht]
    \centering
    \small
    \resizebox{\textwidth}{!}{
        \begin{tabular}{>{\centering\arraybackslash}p{1.2cm} l c c | c c | c c c c c}
            \toprule
            \multirow{2}{*}{Model} & \hspace{6pt}\multirow{2}{*}{Method} & \multirow{2}{*}{Rank} & \multirow{2}{*}{Avg Bits}
            & \multicolumn{2}{c|}{PPL ↓} & \multicolumn{5}{c}{Zero-Shot Accuracy ↑} \\
            \cmidrule(lr){5-6} \cmidrule(lr){7-11}
            & & & & Wiki2 & C4 & Wino & RTE & PiQA & ArcE & ArcC \\
            \midrule
            \multirow{6}{*}{7B}
            & CALDERA  & 64  & 2.1 & 7.34 & \bf{9.50}  & 63.85 & 55.23 & 72.69 & 62.25 & 32.04 \\
            & \cellcolor{Gray}\text{~~~~+ODLRI}  & \cellcolor{Gray}64  & \cellcolor{Gray}2.1 & \cellcolor{Gray}\bf{7.20} & \cellcolor{Gray}9.52 & \cellcolor{Gray}\bf{65.04} & \cellcolor{Gray}\bf{62.45} & \cellcolor{Gray}\bf{72.91} & \cellcolor{Gray}\bf{65.49} & \cellcolor{Gray}\bf{33.70}  \\
            & CALDERA  & 128  & 2.2  & 6.90 & 9.01  & \bf{65.27} & 57.58 & 72.80 & \bf{63.89} & \bf{34.21} \\
            & \cellcolor{Gray}\text{~~~~+ODLRI}  & \cellcolor{Gray}128  & \cellcolor{Gray}2.2 & \cellcolor{Gray}\bf{6.72} & \cellcolor{Gray}\bf{8.82} & \cellcolor{Gray}64.33 & \cellcolor{Gray}\bf{57.76} & \cellcolor{Gray}\bf{75.30} & \cellcolor{Gray}62.63 & \cellcolor{Gray}33.19  \\
            & CALDERA  & 256  & 2.4 & 6.47 & 8.47  & 65.19 & 60.11 & 74.43 & \bf{66.35} & 34.25\\
            & \cellcolor{Gray}\text{~~~~+ODLRI}  & \cellcolor{Gray}256  & \cellcolor{Gray}2.4 & \cellcolor{Gray}\bf{6.33} & \cellcolor{Gray}\bf{8.27} & \cellcolor{Gray}\bf{66.42} & \cellcolor{Gray}\bf{61.01} & \cellcolor{Gray}\bf{74.67} & \cellcolor{Gray}65.49 & \cellcolor{Gray}\bf{36.09} \\
            \midrule
            \multirow{6}{*}{13B}
            & CALDERA  & 64  & 2.08  & 5.92 & \bf{7.96}  & 66.37 & 58.84 & 75.27 & \bf{69.00} & \bf{38.18}  \\
            & \cellcolor{Gray}\text{~~~~+ODLRI}  & \cellcolor{Gray}64  & \cellcolor{Gray}2.08 & \cellcolor{Gray}\bf{5.91} & \cellcolor{Gray}7.97 & \cellcolor{Gray}\bf{67.32} & \cellcolor{Gray}\bf{63.18} & \cellcolor{Gray}\bf{75.90} & \cellcolor{Gray}65.82 & \cellcolor{Gray}37.20  \\
            & CALDERA  & 128  & 2.16  & 5.77 & \bf{7.71}  & 67.4 & \bf{62.45} & 75.70 & \bf{69.65} & 38.91  \\
            & \cellcolor{Gray}\text{~~~~+ODLRI}  & \cellcolor{Gray}128  & \cellcolor{Gray}2.16 & \cellcolor{Gray}\bf{5.73} & \cellcolor{Gray}\bf{7.71}  & \cellcolor{Gray}\bf{69.69} & \cellcolor{Gray}61.37 & \cellcolor{Gray}\bf{76.50} & \cellcolor{Gray}68.69 & \cellcolor{Gray}\bf{39.32}  \\
            & CALDERA  & 256  & 2.32  & 5.56 & 7.39  & \bf{69.13} & \bf{64.08} & 75.98 & 70.26 & 39.88  \\
            & \cellcolor{Gray}\text{~~~~+ODLRI}  & \cellcolor{Gray}256  & \cellcolor{Gray}2.32 & \cellcolor{Gray}\bf{5.46} & \cellcolor{Gray}\bf{7.28}  & \cellcolor{Gray}68.19 & \cellcolor{Gray}62.63 & \cellcolor{Gray}\bf{76.65} & \cellcolor{Gray}\bf{71.48} & \cellcolor{Gray}\bf{39.96}  \\
            \midrule
            \multirow{4}{*}{70B}
            & CALDERA  & 128  & 2.1 & 4.09 & 5.91  & \bf{75.93} & 71.84 & 79.38 & 76.98 & 47.10  \\
            & \cellcolor{Gray}\text{~~~~+ODLRI}  & \cellcolor{Gray}128  & \cellcolor{Gray}2.1 & \cellcolor{Gray}\bf{4.06} & \cellcolor{Gray}\bf{5.89}  & \cellcolor{Gray}75.30 & \cellcolor{Gray}\bf{72.56} & \cellcolor{Gray}\bf{79.98} & \cellcolor{Gray}\bf{78.45} & \cellcolor{Gray}\bf{49.06}  \\
            & CALDERA  & 256  & 2.2  & 3.99 & 5.78  & 75.69 & \bf{72.92} & 80.41 & \bf{78.28} & 49.15 \\
            & \cellcolor{Gray}\text{~~~~+ODLRI}  & \cellcolor{Gray}256  & \cellcolor{Gray}2.2 & \cellcolor{Gray}\bf{3.94} & \cellcolor{Gray}\bf{5.73}  & \cellcolor{Gray}\bf{75.91} & \cellcolor{Gray}70.75 & \cellcolor{Gray}\bf{80.73} & \cellcolor{Gray}77.96 & \cellcolor{Gray}\bf{49.91}  \\
            \specialrule{1.1pt}{1.2pt}{3pt}
            \multicolumn{2}{l}{Uncompressed (7B)} & —  & 16  & 5.12   & 6.63 & 67.3  & 63.2 & 78.5 & 69.3 & 40.0  \\
            \multicolumn{2}{l}{Uncompressed (13B)} & —  & 16  & 4.57   & 6.05 & 69.5  & 61.7 & 78.8 & 73.2 & 45.6  \\
            \multicolumn{2}{l}{Uncompressed (70B)} & —  & 16  & 3.12   & 4.97 & 77.0  & 67.9 & 81.1 & 77.7 & 51.1  \\
            \bottomrule
        \end{tabular}
    }
    \vspace{-0.5em}
    \caption{Comparison of our method with CALDERA on zero-shot perplexities ($\downarrow$) and accuracies ($\uparrow$) of Llama2 models on WikiText-2 and C4. $\bQ$ is quantized 2-Bit and $\bL, \bR$ are quantized 4-Bit. Lower perplexity and higher accuracy values are \textbf{bolded}.}
    \label{tab:zero_shot_evaluation_lr4}

\end{table*}

\paragraph{Models.}
We conduct experiments on a range of large language models, including Llama2 (7B, 13B, and 70B parameters)~\citep{touvron2023llama},
Llama3-8B~\citep{dubey2024llama}, and Mistral-7B~\citep{jiang2023mistral}.

\paragraph{Evaluation Metrics.}
We evaluate quantization performance using perplexity and zero-shot accuracy benchmarks.
Perplexity is measured on the test splits of WikiText-2~\citep{merity2017pointer} and C4~\citep{raffel2020exploring}, 
following each model’s predefined context length (e.g., 4096 tokens for Llama2 and 8192 tokens for Llama3).
Zero-shot accuracy is assessed using the EleutherAI \texttt{lm-evaluation-harness}~\citep{eval-harness}, covering a range of NLP benchmarks, including Winogrande~\citep{sakaguchi2020winogrande}, RTE~\citep{bentivogli2009fifth}, PiQA~\citep{bisk2020piqa}, ARC Easy~\citep{clark2018think}, and ARC Challenge~\citep{clark2018think}.
To ensure statistical reliability, most results are averaged across two independent random seeds
(Details are in~\Cref{app:Additional Experimental Setup}).

\subsection{Effect of LR Initialization on Joint Q+LR Optimization}
We compare three initialization methods for $\bL, \bR$ to evaluate their impact on performance.
The first method, Zero Initialization (0), follows CALDERA’s default setup, where $\bL_0=0$ and $\bR_0=0$.
The second method, LRApprox($\bW$), initializes $\bL, \bR$ using a low-rank approximation (LPLR) of $\bW$.
The third method, ODLRI, is our proposed approach, which explicitly focuses salient weights for low-rank approximation.

\paragraph{Quantization Scale.}
We first examine the quantization scale, which reflects the dynamic range of the weights and directly impacts low-bit quantization efficiency.
A lower quantization scale indicates a more compact weight distribution, enabling finer representation in low-bit precision and reducing overall quantization error.

\Cref{fig:qscale_comp} shows that ODLRI significantly reduces the quantization weight scale, ensuring that weights are mapped into a range more suitable for accurate quantization.
In contrast, baseline methods exhibit higher scales, making quantization more challenging and increasing the risk of performance degradation.
This reduction in quantization scale with ODLRI is a key factor in achieving superior performance in extreme low-bit settings.

Additional results on other layers and different projection types confirm consistent scale reduction by ODLRI (see~\Cref{app: scale and error}).

\paragraph{Activation-aware Error.}
To further validate that ODLRI effectively minimizes our objective, 
we measure the normalized activation-aware error for each layer’s weight component.
This metric evaluates how well the decomposition preserves activation-dependent weight structure:
\begin{align*}
    \frac{\|(\bW - \bQ - \bL \bR)\bX\|^2_\text{F}}{\|\bW\bX\|^2_\text{F}}.
\end{align*}

As illustrated in \Cref{fig:error_comp}, the default CALDERA configuration with zero initialization results in significantly higher activation-aware errors.
Even when using a low-rank approximation of $\bW$, the error remains consistently higher across layers compared to ODLRI.
The substantial reduction in activation-aware error achieved by ODLRI confirms that it represents weight approximations more effectively. Additional results are provided in~\Cref{app: scale and error}.

\subsection{Zero-shot Evaluation}
Following the setup in the previous section,
we conduct zero-shot evaluations of various models with a fixed 2-Bit $\bQ$ component and an $\bL\bR$ component set to either quantized 4-Bit or unquantized 16-Bit.
The goal is to assess how applying ODLRI initialization compares against the baseline CALDERA under different quantization constraints.

\paragraph{4-Bit LR on Llama2.}
\Cref{tab:zero_shot_evaluation_lr4} reports results where the $\bL$ and $\bR$ are quantized to 4-Bit, with LPLR applied for iterative updates.
Despite this additional optimization step, ODLRI consistently improves perplexity (WikiText-2, C4) and zero-shot accuracy across multiple tasks (e.g., PiQA, RTE) for most configurations.
These results highlight that a principled initialization strategy enhances performance, even under aggressive quantization.

Importantly, the only modification from standard CALDERA is the $\bL\bR$ initialization strategy, demonstrating that even in highly constrained quantization settings,
ODLRI retains crucial distributional information that would otherwise can be lost. Additional results under more extreme compression, specifically at lower ranks ($r \leq 32$), can be found in~\Cref{app: lower rank extreme compression}.

\paragraph{16-Bit LR on Llama2.}
\renewcommand{\arraystretch}{1.0}  
\setlength{\tabcolsep}{4pt}  
\definecolor{Gray}{gray}{0.9}
\begin{table}[!ht]
    \centering
    \small
    \resizebox{\columnwidth}{!}{
        \begin{tabular}{>{\centering\arraybackslash}p{0.9cm} l c c >{\centering\arraybackslash}p{0.8cm} >{\centering\arraybackslash}p{0.8cm}}
            \toprule
            \multirow{3}{*}{Model} & \multirow{3}{*}{\hspace{6pt}\centering Method} & \multirow{3}{*}{Rank} & \multirow{3}{*}{Avg Bits}
            & \multicolumn{2}{c}{PPL $\downarrow$} \\
            \cmidrule(lr){5-6}
            & & & & Wiki2 & C4 \\
            \midrule
            \multirow{7}{*}{7B}
            & CALDERA  & 64  & 2.40 & 7.25 & 9.52  \\
            & \cellcolor{Gray}\text{~~~~+ODLRI}  & \cellcolor{Gray}64 & \cellcolor{Gray}2.40 & \cellcolor{Gray}\bf{7.17} & \cellcolor{Gray}\bf{9.41}  \\
            & CALDERA  & 128 & 2.80 & 6.84 & 8.95  \\
            & \cellcolor{Gray}\text{~~~~+ODLRI}  & \cellcolor{Gray}128  & \cellcolor{Gray}2.80 & \cellcolor{Gray}\bf{6.70} & \cellcolor{Gray}\bf{8.79}  \\
            & CALDERA  & 256 & 3.60 & 6.42 & 8.43  \\
            & \cellcolor{Gray}\text{~~~~+ODLRI}  & \cellcolor{Gray}256 & \cellcolor{Gray}3.60 & \cellcolor{Gray}\bf{6.18} & \cellcolor{Gray}\bf{8.23}  \\
            \midrule
            \multirow{7}{*}{13B}
            & CALDERA  & 64 & 2.32 & 5.93 & \bf{7.95}  \\
            & \cellcolor{Gray}\text{~~~~+ODLRI}  & \cellcolor{Gray}64 & \cellcolor{Gray}2.32 & \cellcolor{Gray}\bf{5.90} & \cellcolor{Gray}7.96  \\
            & CALDERA  & 128 & 2.64 & 5.77 & 7.69  \\
            & \cellcolor{Gray}\text{~~~~+ODLRI}  & \cellcolor{Gray}128 & \cellcolor{Gray}2.64 & \cellcolor{Gray}\bf{5.64} & \cellcolor{Gray}\bf{7.54} \\
            & CALDERA  & 256 & 3.28 & 5.48 & 7.32  \\
            & \cellcolor{Gray}\text{~~~~+ODLRI}  & \cellcolor{Gray}256  & \cellcolor{Gray}3.28  & \cellcolor{Gray}\bf{5.38} & \cellcolor{Gray}\bf{7.14}  \\
            \specialrule{1.1pt}{1.2pt}{3pt}
            \multicolumn{2}{l}{Uncompressed (7B)} & — & 16 & 5.12   & 6.63 \\
            \multicolumn{2}{l}{Uncompressed (13B)} & — & 16 & 4.57   & 6.05 \\
            \bottomrule
        \end{tabular}
    }
\vspace{-0.5em}
    \caption{
    Zero-Shot Perplexity ($\downarrow$) of ODLRI with CALDERA for Llama2 models on WikiText-2 and C4. 
    $\bQ$ is 2-Bit quantized, and $\bL, \bR$ are 16-Bit. Lower values are \textbf{bolded}.
    Additional zero-shot accuracy results are provided in \Cref{app: 16bit zeroshot accuracy}.}
%
    \label{tab:zero_shot_perplexity_lr16}
\vspace{-1em}
\end{table}

We further evaluate the models in a 16-Bit $\bL\bR$ setting, where the $\bL\bR$ component is left unquantized (\Cref{tab:zero_shot_perplexity_lr16}).

Without the need for low-bit refinement steps on $\bL\bR$ component, this configuration allows for a direct assessment of ODLRI’s effectiveness.
It enables $\bL_0, \bR_0$ to fully capture the benefits of outlier-driven initialization, preserving fine-grained weight structures that are critical for performance under extreme low-bit settings.
As expected, perplexity and zero-shot accuracy improve compared to the more constrained 4-Bit LR scenario,
providing clearer evidence of ODLRI’s effectiveness without the confounding effects of additional $\bL\bR$ quantization.
As in the 4-Bit $\bL\bR$ setting, ODLRI achieves more pronounced gains, particularly at higher ranks.
The corresponding zero-shot accuracy measurements for 16-Bit LR setting are provided in~\Cref{app: 16bit zeroshot accuracy}.

These results demonstrate that the effectiveness of ODLRI holds across both aggressive 4-Bit $\bL\bR$ settings and relaxed 16-Bit $\bL\bR$ conditions without additional refinement, highlighting its robustness across diverse compression regimes.

\paragraph{4-Bit LR on Llama3-8B and Mistral-7B.}
To assess the generalizability of ODLRI beyond Llama2 models, we evaluate its performance on Llama3-8B and Mistral-7B.
\Cref{tab:llama3_mistral} presents results for these models with 4-Bit quantization applied to the $\bL\bR$ component.
Under a range of rank configurations, ODLRI consistently improves over original CALDERA by more effectively capturing salient weight directions, resulting in lower perplexity on both WikiText-2 and C4.
These results demonstrate that the effectiveness of ODLRI generalizes beyond Llama2 models.
We also provide results on non-LLaMA models and alternative quantizers in~\cref{app:ODLRI_nonLlama}.
\renewcommand{\arraystretch}{0.8} 

\begin{table}[!ht]
    \begin{center}     
        \resizebox{\columnwidth}{!}{
            \begin{tabular}{@{}clcccc@{}}
                \toprule
                \multirow{2}{*}{Rank} & \hspace{8pt}\multirow{2}{*}{Method} & \multicolumn{2}{c}{Llama3-8B} & \multicolumn{2}{c}{Mistral-7B} \\ 
                \cmidrule(l){3-6}
                & & Wiki2 ↓ & C4 ↓ & Wiki2 ↓ & C4 ↓  \\ 
                \midrule
                \multirow{2}{*}{64}                      
                & CALDERA & 10.58 & 11.35 & \bf{6.37} & 7.11 \\
                &\cellcolor{Gray}\text{~~~~+ODLRI} & \cellcolor{Gray}\bf{10.35} & \cellcolor{Gray}\bf{11.15} & \cellcolor{Gray}\bf{6.37} & \cellcolor{Gray}\bf{7.10} \\
                \midrule
                \multirow{2}{*}{128}                      
                & CALDERA & 9.41 & \bf{10.21} & 6.11 & 6.89 \\
                &\cellcolor{Gray}\text{~~~~+ODLRI} & \cellcolor{Gray}\bf{9.35} & \cellcolor{Gray}10.32 & \cellcolor{Gray}\bf{6.08} & \cellcolor{Gray}\bf{6.86} \\
                \midrule
                \multirow{2}{*}{256}                     
                & CALDERA & 8.70 & 9.77 & 5.77 & 6.59 \\
                &\cellcolor{Gray}\text{~~~~+ODLRI} & \cellcolor{Gray}\bf{8.12} & \cellcolor{Gray}\bf{9.33} & \cellcolor{Gray}\bf{5.69} & \cellcolor{Gray}\bf{6.53} \\
                \bottomrule
            \end{tabular}
        }
    \end{center}
\vspace{-0.5em}
    \caption{
    Zero-Shot Perplexity ($\downarrow$) of ODLRI with CALDERA for Llama3-8B and Mistral-7B on WikiText-2 and C4. 
    $\bQ$ is 2-Bit quantized, while $\bL, \bR$ are 4-Bit quantized. Lower values are \textbf{bolded}.}
    \label{tab:llama3_mistral}
\end{table}

\subsection{Number of Outlier Columns ($k$)}
In ODLRI, we determine salient weight components by explicitly targeting activation outliers, selecting top-$k$ activations corresponding to outlier-sensitive weights.
Instead of following the rank-based selection that picks the top-$r$ components based solely on low-rank dimension $r$,
we set $k<r$ to intensively focus on outliers.

\renewcommand{\arraystretch}{0.8} 
\definecolor{Gray}{gray}{0.9}
\begin{table}[!ht]
    \begin{center}
    \footnotesize
        \resizebox{0.9\columnwidth}{!}{
            \begin{tabular}{@{}ccccc@{}}
                \toprule
                \multirow{2}{*}{ODLRI} & \multicolumn{2}{c}{$\bL$, $\bR$ 16-Bit} & \multicolumn{2}{c}{$\bL$, $\bR$ 4-Bit} \\ 
                \cmidrule(l){2-5}
                & Wiki2~↓ & C4~↓ & Wiki2~↓ & C4~↓  \\ 
                \midrule                   
                $\bH_o$ $(k=r)$ & 6.38 & 8.43 & 6.46 & 8.52 \\
                \rowcolor{Gray}$\bH_o$ $(k<r)$& {\bf 6.18} & {\bf 8.23} & {\bf 6.33} & {\bf 8.27} \\
                \bottomrule
            \end{tabular}
        }
    \end{center}
\vspace{-0.5em}
    \caption{Comparison of OLDRI with various values of \( k \), specifically \( \mathbf{H_o}~(k = 256) \) and \( \mathbf{H_o}~(k = 16) \) by perplexities (\(\downarrow\)) of Llama2-7B on WikiText-2 and C4. \(\bQ\) is 2-Bit, and $\bL$, $\bR$ are either 16-Bit or quantized 4-Bit with a rank of 256. Lower perplexity values are \textbf{bolded}. Details regarding selection of $k$ are provided in~\Cref{app: Rank-Dependent Outlier Selection}.
}
    \label{tab:table_odlri_hessian_ppl}
\end{table}

\vspace{-0.5em}
To evaluate the impact of this selection, we measure perplexity (PPL) under different configurations.
Our results show that choosing $k$ based on activation outliers consistently outperforms selecting the top-$r$ components through standard low-rank approximation.
This improvement highlights the importance of leveraging activation statistics to guide low-rank approximation,
ensuring that the most outlier-sensitive weights are efficiently modeled in $\bL\bR$.
These findings validate that ODLRI’s targeted selection enhances low-rank approximation, leading to better quantization performance in extreme low-bit settings.

\section{Conclusion}
In this work, we investigated the role of initialization strategies in iterative joint optimization approaches for LLM compression.
We introduce a unified framework that reformulates joint optimization algorithms through the lens of low-rank component initialization.
Our analysis shows that the choice of initialization determines the entire trajectory of weight decomposition by assigning persistent roles to the components. Based on these insights, we proposed ODLRI, which assigns a distinct role to the low-rank component to capture activation-sensitive weights while using quantization for the remaining weights.
Through extensive experiments across various LLM architectures, we demonstrated that incorporating ODLRI significantly improves model performance and compression stability. Consequently, our approach advances the practical implementation of efficient LLM compression and steers us toward optimal weight decomposition.

\section*{Acknowledgement}
This work was supported by Institute of Information \& communications Technology 
Planning \& Evaluation (IITP) grant funded by the Korea government(MSIT) 
(No. RS-2024-00457882, National AI Research Lab Project),
IITP grant funded by the Korean Government (MSIT)
(No. RS-2020-II201361, Artificial Intelligence Graduate School Program (Yonsei University)),
and K-CHIPS(Korea Collaborative \& High-tech Initiative for Prospective Semiconductor Research)
(RS-2024-00405946, 24052-15TC) funded by the Ministry of Trade, Industry \& Energy (MOTIE, Korea).

\section*{Limitations}
Our work focuses on weight-only quantization, optimizing the decomposition of model weights into quantized and low-rank components.
While this approach enhances low-bit quantization performance, it does not address the quantization of activations or KV cache, which are critical for further improving inference efficiency.
Joint weight and activation quantization introduces additional challenges, such as distributional shifts and increased sensitivity to outliers, requiring specialized calibration techniques.
Similarly, KV cache quantization is essential for reducing memory overhead in long-context inference but remains outside the scope of this study.

Additionally, while we evaluate ODLRI within CALDERA, we believe that our approach can be applied to other joint $\bQ+\bL\bR$ optimization algorithms beyond CALDERA.
Exploring how ODLRI integrates with different iterative quantization frameworks presents an interesting direction for future research.

\section*{Ethical Considerations}
We focus on efficient compression of LLMs through quantization and low-rank decomposition. While these techniques improve computational efficiency and enable broader deployment, they also present ethical considerations.

First, compressed models may inherit and amplify biases present in the original model. Ensuring fairness and preventing unintended distortions due to quantization artifacts remain critical challenges.

Second, model compression enhances accessibility but may also facilitate misuse, including deployment in applications without adequate ethical oversight. Responsible usage and regulation are essential.

We emphasize the importance of ethical AI development and encourage continued evaluation of the societal impact of model compression.

\bibliography{custom}

\clearpage
\appendix

\section{Additional Experimental Setup}
\label{app:Additional Experimental Setup}
\paragraph{Calibration Data and Hessian Computation.}
We use precomputed Hessians from \citet{Hessians-Llama-2} for Llama2 models.
For Llama3-8B and Mistral-7B, Hessians are computed directly to ensure alignment with model-specific architectures.
Hessians are computed using 256 randomly sampled examples from the RedPajama dataset~\citep{weber2024redpajama}.
Context lengths are set to 4096 tokens for Llama2 and 8192 tokens for Llama3, maintaining consistency with each model’s native configuration.

\begin{table*}[!ht]
\centering
\renewcommand{\arraystretch}{1.1}
\resizebox{0.9\textwidth}{!}{%
\begin{tabular}{l l c l}
\toprule
\hspace{20pt}\textbf{Model} & \hspace{30pt}\textbf{Source} & \textbf{Accessed via} & \hspace{55pt}\textbf{License} \\
\midrule
Llama2-7B & \cite{touvron2023llama} & \href{https://huggingface.co/meta-llama/Llama-2-7b}{Link} & Llama 2 Community License \\
Llama2-13B & \cite{touvron2023llama} & \href{https://huggingface.co/meta-llama/Llama-2-13b}{Link} & Llama 2 Community License \\
Llama2-70B & \cite{touvron2023llama} & \href{https://huggingface.co/meta-llama/Llama-2-70b}{Link} & Llama 2 Community License \\
Llama3-8B & \cite{dubey2024llama} & \href{https://huggingface.co/meta-llama/Meta-Llama-3-8B}{Link} & Meta Llama 3 Community License \\
Mistral-7Bv0.1 & \cite{jiang2023mistral} & \href{https://huggingface.co/mistralai/Mistral-7B-v0.1}{Link} & Apache license 2.0 \\
\bottomrule
\end{tabular}%
\label{tab:model_license}
}
\caption{Summary of models used in this paper.} 
\label{tab:models}
\end{table*}

\begin{table*}[!ht]
\centering
\renewcommand{\arraystretch}{1.1}
\resizebox{0.9\textwidth}{!}{%
\begin{tabular}{l l c l}
\toprule
\hspace{40pt}\textbf{Dataset} & \hspace{25pt}\textbf{Source} & \textbf{Accessed via} & \hspace{25pt}\textbf{License} \\
\midrule
RedPajama-Data-1T-Sample & \cite{weber2024redpajama} & \href{https://huggingface.co/datasets/togethercomputer/RedPajama-Data-1T-Sample}{Link} & Apache License 2.0\\
Hessians-Llama-2-7b-6144 & \cite{Hessians-Llama-2} & \href{https://huggingface.co/relaxml/Hessians-Llama-2-7b-6144}{Link} & - \\
Hessians-Llama-2-13b-6144 & \cite{Hessians-Llama-2} & \href{https://huggingface.co/relaxml/Hessians-Llama-2-13b-6144}{Link} & - \\
Hessians-Llama-2-70b-6144 & \cite{Hessians-Llama-2} & \href{https://huggingface.co/relaxml/Hessians-Llama-2-70b-6144}{Link} & - \\
Wikitext-2-raw-v1 & \cite{merity2017pointer} & \href{https://huggingface.co/datasets/Salesforce/wikitext/tree/main/wikitext-2-raw-v1}{Link} & CC-BY-SA-3.0 \\
C4 & \cite{raffel2020exploring} & \href{https://huggingface.co/datasets/allenai/c4}{Link} & ODC-BY \\
lm-eval-harness & \cite{eval-harness} & \href{https://github.com/EleutherAI/lm-evaluation-harness}{Link} & MIT License \\
Winogrande & \cite{sakaguchi2020winogrande} & \href{https://huggingface.co/datasets/allenai/winogrande}{Link} & Apache License 2.0 \\
RTE & \cite{bentivogli2009fifth} & \href{https://huggingface.co/datasets/nyu-mll/glue}{Link} & Apache License 2.0 \\
PiQA & \cite{bisk2020piqa} & \href{https://huggingface.co/datasets/ybisk/piqa}{Link} & Apache License 2.0 \\
ARC Easy & \cite{clark2018think} & \href{https://huggingface.co/datasets/allenai/ai2_arc}{Link} & CC-BY-SA-4.0 \\
ARC Challenge & \cite{clark2018think} & \href{https://huggingface.co/datasets/allenai/ai2_arc}{Link} & CC-BY-SA-4.0 \\
\bottomrule
\end{tabular}%
\label{tab:dataset_license}
}
\caption{Summary of datasets used in this paper.} 
\label{tab:datasets}
\end{table*}

\paragraph{Hardware Environment.}
Experiments were conducted on both consumer- and enterprise-grade GPUs. 
The primary pipeline was executed on NVIDIA GeForce RTX 3090, RTX 4090, and RTX A6000 GPUs, while the quantization and evaluation of the Llama2-70B model were performed on NVIDIA L40s GPU clusters due to its large size.

Computation time for quantization and evaluation varied depending on the model size and hardware configuration. For Llama2-7B and Mistral-7B, quantization was completed in 4 GPU hours on a cluster of six GPUs. 
For Llama2-13B, quantization required 8 GPU hours on a six-GPU cluster. 
Besides, for Llama3-8B, quantization was achieved in 5 GPU hours on a cluster of six GPUs. 
For Llama2-70B, quantization was conducted using four NVIDIA L40S GPUs, completing within 48 hours. In the absence of parallel processing, the estimated runtime would have increased proportionally to the number of GPUs used. Evaluation phase was carried out on the same GPUs utilized for quantization. Perplexity measurement required approximately 0.5 GPU hours, while zero-shot evaluation consumed around 1.5 GPU hours. Since pretrained models were used, no additional training time was required.

All experiments were repeated using two different random seeds, with the exception of Llama2-70B, for which only one seed was used due to resource constraints. Overall, the quantization and evaluation processes accumulated approximately 4000 GPU hours in total.

\paragraph{CALDERA's Default Configuration.}
In our experiments, the CALDERA model was configured with the following settings: \verb|hadamard_transform| was set to \verb|true|, \verb|outer_iter| to \verb|15|, \verb|inner_iter| to \verb|10|, \verb|rand_svd| to \verb|false|, \verb|Q_hessian_downdate| to \verb|false|, and \verb|update_order| to \verb|Q LR|.

\paragraph{Summary of Model and Dataset.}
We present a summary of the models and datasets employed in this paper. The detailed specifications, including sources, access methods, and licensing terms, are summarized in Table~\ref{tab:models}. Also, the datasets and their corresponding metadata are detailed in Table~\ref{tab:datasets}.

\section{Outlier-Aware Initialization Detail}
\subsection{ODLRI Method in Detail} \label{app: Outlier sensitive Weight SVD for LR Initialization}
To process outlier‐sensitive weights by low-rank approximation, we decompose our optimization objective 
$$
\mathbf{W}\mathbf{X} = \mathbf{W}(\bX_{\text{o}}+ \bX_{\text{r}}) = \mathbf{W}\bX_{\text{o}}+ \mathbf{W}\bX_{\text{r}}
$$
into two distinct components. 
Here, $\bX_{\text{o}}$ represents the channels containing outliers (with other channels set to zero), while $\bX_{\text{r}}$ indicates the remaining channels (with outlier channels zeroed). 
Since the non-zero entries of $\bX_{\text{o}}$ and $\bX_{\text{r}}$ are mutually exclusive along the channel dimension, both $\bX_{\text{o}}$ and $\bX_{\text{r}}$ retain the original matrix dimensions. 

Standard activation-aware low-rank factorization, which solves
\begin{align*}
\argmin_{\bL, \bR} \|(\bW-\bL\bR)\bX\|^2_\text{F}.
\end{align*}
Using the empirical second-moment matrix $\bH = \bX\bX^\top$, this objective is equivalent to:
\begin{align*}
\argmin_{\bL, \bR} \|(\bW-\bL\bR)\bH(\bW-\bL\bR)^\top\|.
\end{align*}

However, this formulation treats all activations equally, leading to a low-rank approximation that does not explicitly focus on outliers,
which are often the primary bottleneck in quantization.

To ensure that the low-rank component captures only the most challenging weight structures, we define a restricted activation covariance matrix $\bH_{\text{o}}$,
a submatrix of $\bH$ that prioritizes outlier-sensitive channels:
\begin{equation}
(\mathbf{H}_{o})_{ij} =
\begin{cases}
\mathbf{H}_{ij}, & \text{if } i,j \in \mathcal{I}, \\
0, & \text{otherwise}
\end{cases}
\end{equation}
where the subset of indices \(\mathcal{I} \subset \{1, \ldots, d\}\) corresponds to the top-$k$ ($k$ < $r$) channels with the highest diagonal values of $\bH$.
These channels correspond to the most dominant activation patterns, which often align with weight outliers that distort quantization performance.

We then solve the outlier-aware optimization problem:
\begin{align*}
\bL_0, \bR_0 =\argmin_{\bL, \bR} \|(\bW-\bL\bR)\bH_{\text{o}}(\bW-\bL\bR)^\top\|.
\end{align*}

This is achieved by applying a Cholesky decomposition to the selected Hessian submatrix $\bH_{\text{o}}$, which encodes the activation covariance information of high-variance channels:
\begin{align*} 
    \bH_{\text{o}} = \bS_{\text{o}}\bS_{\text{o}}^\top, 
\end{align*}
where $\bS_{\text{o}}$ is a lower triangular matrix.

Unlike SVD-LLM~\citep{wang2025svdllm}, which applies data whitening to the entire Hessian matrix \(\mathbf{H}\), our transformation performs selective whitening on the outlier subset \(\mathbf{H}_\text{o}\) corresponding to $\bX_{\text{o}}$. This selective whitening improves the numerical conditioning for the subsequent SVD while ensuring that the low-rank component effectively captures the salient weight information.

Once the whitening transformation is applied, we perform SVD on the transformed weight matrix $\bW\bS_{\text{o}}$,
truncating the decomposition to rank $r$ to obtain the outlier-focused low-rank components:
\begin{align*}
    \text{SVD}(\bW\bS_{\text{o}}) = \mathbf{U}_{:,:r} \mathbf{\Sigma}_{:r,:r} \mathbf{V}_{:r,:}^\top. 
\end{align*}

Here, the Python-style slicing notation \(\mathbf{U}_{:,:r}\) indicates that we select all rows and the first \(r\) columns of \(\mathbf{U}\). Similarly, \(\mathbf{\Sigma}_{:r,:r}\) denotes the upper-left \(r \times r\) block of \(\mathbf{\Sigma}\), and \(\mathbf{V}_{:r,:}\) selects the first \(r\) rows of \(\mathbf{V}\).

To ensure that the decomposition remains truncation-aware, we initialize the low-rank and quantization components as follows:
\begin{align*} 
\bL_0 =& \mathbf{U}_{:,:r} \sqrt{\mathbf{\Sigma}_{:r,:r}}\\
\bR_0 =& \sqrt{\mathbf{\Sigma}_{:r,:r}} \mathbf{V}_{:r,:}^\top \mathbf{S}_{\text{o}}^{-1}. 
\end{align*}

By ensuring that the residual weight matrix $\bW-\bL_0\bR_0$ has been preconditioned to remove outlier effects,
we significantly reduce quantization error compared to conventional activation-aware quantization approaches.
This allows the quantized component 
$\bQ$ to operate on a more uniform weight distribution, improving numerical stability:
\begin{align*} 
\mathbf{Q}_1 = \text{Quantize} (\mathbf{W} - \mathbf{L}_0 \mathbf{R}_0).
\end{align*}

\subsection{Rank-Dependent Outlier Selection of $k$} \label{app: Rank-Dependent Outlier Selection}
In our experiments, we denote the rank of the low-rank components \(\mathbf{L}\) and \(\mathbf{R}\) as \(r\), and we set the number of outlier-sensitive columns \(k\) in proportion to this rank. Specifically, we define:
\[
k = p \times n,
\]
where \(n\) is the dimension of the calibration Hessian \(\mathbf{H} \in \mathbb{R}^{n \times n}\) and \(p\) is a rank-dependent percentage.

We adopt the following settings:
\begin{itemize}
    \item For \(r = 64\): \(p = 0.1\%\)
    \item For \(r = 128\): \(p = 0.2\%\)
    \item For \(r = 256\): \(p = 0.4\%\)
\end{itemize}

For example, consider the key projection matrix of Llama2-7B. Its corresponding Hessian matrix has a shape of \(4096 \times 4096\). When we set the rank \(r = 256\), we use \(p = 0.4\%\). Thus, the number of outlier-sensitive columns is computed as:
\[
k = p \times n = 0.4\% \times 4096 \approx 16.
\]
This means that in this example, 16 outlier-sensitive columns are selected for the key projection.

\subsection{ODLRI's Impact on Salient Weights}
\renewcommand{\arraystretch}{1.3} 
\begin{table}[!ht]
    \vspace{-1em}
    \begin{center}    
    \small
        \resizebox{\columnwidth}{!}{
            \begin{tabular}{@{}ccccc@{}}
                \toprule
                \rule{0pt}{3ex}
                \large{Hessian} & $\displaystyle\frac{\|\bL\bR\bX_{\text{o}}\|}{\|\bW\bX_{\text{o}}\|}$ & $\displaystyle\frac{\|\mathbf{E}_{LR}\bX_{\text{o}}\|}{\|\bW\bX_{\text{o}}\|}$ & $\displaystyle\frac{\|\bL\bR\bX_r\|}{\|\bW\bX_r\|}$ & $\displaystyle\frac{\|\mathbf{E}_{LR}\bX_r\|}{\|\bW\bX_r\|}$  \\ 
                \midrule                
                {$\bH$} & 0.997 & 0.073 & 0.920 & 0.392 \\
                \midrule
                \rowcolor{Gray}{$\bH_{o}$} & 0.999 & 0.001 & 0.903 & 0.430 \ \\
                \bottomrule
            \end{tabular}
        }
    \end{center}
    \vspace{-0.5em}
    \caption{Effect of Hessian selections in ODLRI. Results are shown for Layer 10 Key Projection matrix of Llama2-7B when using ODLRI initialization. We compare normalized norm of $\|\bX_o\|$ and $\|\bX_r\|$ for $\bL\bR$ and $\mathbf{E}_{LR}$ when using ODLRI initialization.  L2-norm is denoted by $\|\cdot\|$. And $\mathbf{E}_{LR} = \bW - \bL\bR$.}
    \label{tab:table_outlier_sep}
\end{table}
 
This section validates that ODLRI effectively captures and preserves salient weights by explicitly targeting outlier activations in $\bX$.
Unlike standard low-rank approximations that distribute capacity across all activations, ODLRI selectively focuses on outlier activations,
ensuring that the low-rank component is structured to represent the most outlier-sensitive weights before quantization.

To demonstrate this, we compare two hessian formulations for guiding the low-rank approximation.
The first uses the full activation matrix $\bH=\bX\bX^\top$, while the second employs only the top-$k$ outlier activations, $\bX_{\text{o}}$, 
forming the restricted hessian matrix $\bH_{\text{o}} = \bX_{\text{o}}\bX_{\text{o}}^\top$.
The effectiveness of each approach is assessed by evaluating the low-rank representation $\bL\bR$ through the activation norm $\|\bL\bR\bX_{\text{o}}\|$
compared to $\|\bW\bX_{\text{o}}\|$.

\Cref{tab:table_outlier_sep} presents that using $\bH_{\text{o}}$ yields a significantly closer approximation of $\bW\bX_{\text{o}}$ than using $\bH$,
confirming that ODLRI’s outlier-driven initialization better preserves salient weights.
Moreover, the approximation for non-salient weights ($\bX_{\text{r}}$) remains stable,
demonstrating that prioritizing activation outliers does not degrade overall representation.

\section{Additional Results}
\renewcommand{\arraystretch}{0.85}  
\setlength{\tabcolsep}{3.5pt}  
\definecolor{Gray}{gray}{0.9}
\begin{table*}[!t]
    \centering
    \footnotesize
    \resizebox{0.65\textwidth}{!}{
        \begin{tabular}{>{\centering\arraybackslash}p{1.1cm} l c | c c c c c}
            \toprule
            \multirow{2}{*}{Model} & \hspace{7pt}\multirow{2}{*}{Method} & \multirow{2}{*}{Rank} 
            & \multicolumn{5}{c}{Zero-Shot Accuracy ↑} \\
            \cmidrule(lr){4-8}
            & & & Wino & RTE & PiQA & ArcE & ArcC \\
            \midrule
            \multirow{7}{*}{7B}
            & CALDERA  & 64  & 63.73 & \bf{55.59} & \bf{73.4} & 61.93 & 31.27 \\
            & \cellcolor{Gray}\text{~~~~+ODLRI}  & \cellcolor{Gray}64  & \cellcolor{Gray}\bf{65.75} & \cellcolor{Gray}55.23 & \cellcolor{Gray}72.91 & \cellcolor{Gray}\bf{64.44} & \cellcolor{Gray}\bf{31.74}  \\
            & CALDERA  & 128  & \bf{63.97} & \bf{59.03} & 73.53 & 64.58 & 32.94  \\
             & \cellcolor{Gray}\text{~~~~+ODLRI}  & \cellcolor{Gray}128  & \cellcolor{Gray}63.93 & \cellcolor{Gray}\bf{59.03} & \cellcolor{Gray}\bf{73.64} & \cellcolor{Gray}\bf{64.86} & \cellcolor{Gray}\bf{33.75}  \\
            & CALDERA  & 256  & \bf{66.1} & 60.47 & \bf{74.45} & 64.62 & 34.00  \\
            & \cellcolor{Gray}\text{~~~~+ODLRI}  & \cellcolor{Gray}256  & \cellcolor{Gray}64.57 & \cellcolor{Gray}\bf{61.46} & \cellcolor{Gray}65.12 & \cellcolor{Gray}\bf{71.58} & \cellcolor{Gray}\bf{36.31}  \\
            \cmidrule(lr){2-8}
            & QuIP\# & 0 & 61.7 & 57.8 & 69.6 & 61.2 & 29.9 \\
            \midrule
            \multirow{7}{*}{13B}
            & CALDERA  & 64  & 67.36 & 58.84 & 75.07 & 68.11 & \bf{37.58}  \\
            & \cellcolor{Gray}\text{~~~~+ODLRI}  & \cellcolor{Gray}64  & \cellcolor{Gray}\bf{69.85} & \cellcolor{Gray}\bf{67.51} & \cellcolor{Gray}\bf{75.41} & \cellcolor{Gray}\bf{69.91} & \cellcolor{Gray}37.37  \\
            & CALDERA  & 128  & \bf{69.85} & 64.98 & 75.9 & \bf{70.75} & 38.82  \\
            & \cellcolor{Gray}\text{~~~~+ODLRI}  & \cellcolor{Gray}128  & \cellcolor{Gray}68.98 & \cellcolor{Gray}\bf{65.83} & \cellcolor{Gray}\bf{76.14} & \cellcolor{Gray}69.59  &  \cellcolor{Gray}\bf{39.93} \\
            & CALDERA  & 256  & 67.12 & 59.02 & 76.24 & 70.28 & 39.08  \\
            & \cellcolor{Gray}\text{~~~~+ODLRI}  & \cellcolor{Gray}256  & \cellcolor{Gray}\bf{70.40} & \cellcolor{Gray}\bf{67.15} & \cellcolor{Gray}\bf{76.55} & \cellcolor{Gray}\bf{71.63} & \cellcolor{Gray}\bf{41.04}  \\
            \cmidrule(lr){2-8}
            & QuIP\# & 0 & 63.6 & 54.5 & 74.2 & 68.7 & 36.2 \\
            \specialrule{1.1pt}{1.2pt}{3pt}
            \multicolumn{2}{l}{Uncompressed (7B)} & — & 67.3  & 63.2 & 78.5 & 69.3 & 40.0  \\
            \multicolumn{2}{l}{Uncompressed (13B)} & — & 69.5  & 61.7 & 78.8 & 73.2 & 45.6  \\
            \bottomrule
        \end{tabular}
    }
    \vspace{-0.5em}
    \caption{Comparison of our method with CALDERA by zero-shot accuracies ($\uparrow$) of Llama2 models. $\bQ$ is quantized 2-Bit and $\bL, \bR$ are 16-Bit. Higher accuracy values are are \textbf{bolded}.}
    \label{tab:zero_shot_zsa_lr16}
\end{table*}

\subsection{16-Bit LR on Llama2} \label{app: 16bit zeroshot accuracy}
\Cref{tab:zero_shot_zsa_lr16} presents the results of zero-shot accuracy under the setting of 2-Bit $\bQ$ and 16-Bit $\bL\bR$. 
Overall, the results show that the method incorporating ODLRI outperforms the baseline CALDERA approach.

\renewcommand{\arraystretch}{0.8}  
\setlength{\tabcolsep}{3pt}  
\definecolor{Gray}{gray}{0.9}
\begin{table*}[ht]
    \centering
    \small
    \resizebox{0.8\textwidth}{!}{
        \begin{tabular}{>{\centering\arraybackslash}p{1.2cm} l c | c c | c c c c c c}
            \toprule
            \multirow{2}{*}{Rank} & \hspace{6pt}\multirow{2}{*}{Method}  & \multirow{2}{*}{Avg Bits}
            & \multicolumn{2}{c|}{PPL ↓} & \multicolumn{5}{c}{Zero-Shot Accuracy ↑} \\
            \cmidrule(lr){4-5} \cmidrule(lr){6-10}
            & & & Wiki2 & C4 & Wino & RTE & PiQA & ArcE & ArcC \\
            \midrule
            \multirow{2}{*}{16}
            & CALDERA  & 2.025 & 7.88 & 10.16  & \bf{62.64} & 51.26 & 72.09 & \bf{60.63} & \bf{31.48} \\
            & \cellcolor{Gray}\text{~~~~+ODLRI}   & \cellcolor{Gray}2.025 & \cellcolor{Gray}\bf{7.79} & \cellcolor{Gray}\bf{10.02} & \cellcolor{Gray}61.8 & \cellcolor{Gray}\bf{59.57} & \cellcolor{Gray}\bf{72.36} & \cellcolor{Gray}59.43 & \cellcolor{Gray}30.46  \\
            \midrule
            \multirow{2}{*}{32}
            & CALDERA  & 2.05  & 7.85 & 10.18  & 62.9 & 56.31 & 71.7 & 59.00 & 29.52  \\
            & \cellcolor{Gray}\text{~~~~+ODLRI}   & \cellcolor{Gray}2.05 & \cellcolor{Gray}\bf{7.47} & \cellcolor{Gray}\bf{9.75} & \cellcolor{Gray}\bf{65.03} & \cellcolor{Gray}\bf{62.45} & \cellcolor{Gray}\bf{72.52} & \cellcolor{Gray}\bf{62.28} & \cellcolor{Gray}\bf{31.65}  \\
            \bottomrule
        \end{tabular}
    }
    \vspace{-0.5em}
    \caption{Comparison of our method with CALDERA on zero-shot perplexities ($\downarrow$) and accuracies ($\uparrow$) of Llama2 models on extreme low bit setting. $\bQ$ is quantized 2-Bit and $\bL, \bR$ are quantized 4-Bit. Lower perplexity and higher accuracy values are \textbf{bolded}.}
    \label{tab:lower_rank_lr4}

\end{table*}

\subsection{Lower Rank, Extreme Compression on Llama2} \label{app: lower rank extreme compression}
To further evaluate the robustness of our method under more aggressive compression settings, we conduct additional experiments at significantly lower ranks, specifically $r=16$ and $r=32$. 
These settings simulate extreme compression scenarios, where the capacity of the low-rank residual becomes severely constrained. \\
We evaluate on the Llama2-7B model using the same setup as our main experiments: $\mathbf{Q}$ is quantized to 2 bits, and $\mathbf{L}, \mathbf{R}$ to 4 bits. \Cref{tab:lower_rank_lr4} reports perplexity and accuracy across a range of zero-shot evaluation benchmarks. \\
Despite the reduced rank and further limited representational capacity, our proposed ODLRI initialization continues to yield improvements over CALDERA. 
These findings confirm that ODLRI remains effective under severe rank constraints, preserving performance.

\subsection{ODLRI on non-Llama Models and Alternative Quantizers.}\label{app:ODLRI_nonLlama}
\renewcommand{\arraystretch}{1.1} 
\definecolor{Gray}{gray}{0.9}
\begin{table}[!ht]
    \begin{center}
    \footnotesize
        \resizebox{0.9\columnwidth}{!}{
            \begin{tabular}{@{}c|cc|cc@{}}
                \toprule
                \multirow{2}{*}{Method} & \multicolumn{2}{c|}{Llama2-7B} & \multicolumn{2}{c}{Gemma2-2B} \\ 
                \cmidrule(l){2-5}
                & $r=32$ & $r=64$ & $r=32$ & $r=64$  \\ 
                \midrule       
                FP16 & \multicolumn{2}{c|}{8.71} & \multicolumn{2}{c}{13.08} \\
                MXINT-base & 10.62 & 10.38 & 18.72 & 17.65 \\
                \cellcolor{Gray}{\text{~~~~+ODLRI}} & \cellcolor{Gray}{\textbf{10.57}} & \cellcolor{Gray}{\textbf{10.26}} & \cellcolor{Gray}{\textbf{18.52}} & \cellcolor{Gray}{\textbf{17.17}} \\
                \bottomrule
            \end{tabular}
        }
    \end{center}
\vspace{-0.5em}
    \caption{Comparison of OLDRI with MXINT-base on zero-shot perplexities ($\downarrow$) of Llama2-7B and Gemma2-2B on low bit setting. $\bQ$ is quantized 3-Bit and $\bL$,$\bR$ are 16-Bit. Lower perplexity values are \textbf{bolded}.
}
    \label{tab:table_odlri_qera}
\end{table}

Abnormal outliers have been observed in Llama-style LLMs~\citep{yang2024mitigating, yu2024super}, including the Llama family and Mistral, possibly due to their model architecture. 
To assess whether our outlier-aware method remains effective beyond this setting, we evaluate the robustness of ODLRI in two distinct scenarios: (1) non-Llama architectures and (2) alternative quantization methods. 
Specifically, we conduct experiments on both Gemma2-2B~\citep{riviere2024gemma} and Llama2-7B using the MXINT~\citep{darvish2023shared} quantizer. 
We replace QuIP\#~\citep{tsengquip} with MXINT (3-bit, block size 32) for both models.

We define MXINT-base as a baseline that follows the same quantize-then-approximate structure as CALDERA: it first applies MXINT quantization and then performs low-rank approximation using the same activation-aware SVD used in CALDERA, without any outlier-driven refinement.

Gemma2-2B serves to test whether ODLRI generalizes to architectures outside the Llama family. In contrast, Llama2-7B is included to isolate and examine the impact of changing the quantizer while keeping the model architecture fixed. For both models, we adopt the same activation-aware low-rank approximation method as CALDERA and perform 15 outer iterations alternating between quantization and low-rank refinement.

Perplexity is measured using \texttt{lm-eval-harness}, which may yield different values from those in the main text. 
Hence, we also report FP16 perplexities for reference.
We note that the $\bL\bR$ component is kept in 16-Bit precision for all experiments.
As shown in~\cref{tab:table_odlri_qera}, ODLRI consistently reduces perplexity compared to MXINT-base. 
These findings confirm that ODLRI remains effective across both non-Llama models and alternative quantization schemes.

\subsection{Effect of LR Initialization in CALDERA} \label{app: Effect of LR initialization}
\renewcommand{\arraystretch}{1.2}  
\setlength{\tabcolsep}{5pt}  
\begin{table}[!ht]
    \centering
    \Large
    \resizebox{\columnwidth}{!}{%
    \begin{tabular}{c | c | cc | cc}
        \toprule
        \multirow{3}{*}[-0.5em]{\centering Weight Type} & 
        \multirow{3}{*}[-0.5em]{\centering Iteration} & 
        \multicolumn{4}{c}{Initialization Method} \\
        \cmidrule(lr){3-6}
         & & \multicolumn{2}{c|}{\textbf{0}}  
           & \multicolumn{2}{c}{LRApprox($\mathbf{W}$)} \\  
        \cmidrule(lr){3-4} \cmidrule(lr){5-6}
         & & $\|\bQ\bX\|$ & $\|\bL\bR\bX\|$
           & $\|\bQ\bX\|$ & $\|\bL\bR\bX\|$ \\
        \midrule
        \multirow{2}{*}{Key Proj.}    & First  & 0.999  & 0.014  & 0.158 & 0.915 \\
                                      & Last   & 0.961  & 0.073  & 0.401 & 0.664 \\
        \midrule
        \multirow{2}{*}{Query Proj.}  & First  & 0.999  & 0.014  & 0.148 & 0.924 \\
                                      & Last   & 0.956  & 0.073  & 0.408 & 0.657 \\
        \midrule
        \multirow{2}{*}{Value Proj.}  & First  & 0.993  & 0.072  & 0.378 & 0.885 \\
                                      & Last   & 0.970  & 0.264  & 0.622 & 0.676 \\
        \midrule
        \multirow{2}{*}{O Proj.}      & First  & 0.993  & 0.038  & 0.373 & 0.886 \\
                                      & Last   & 0.958  & 0.149  & 0.495 & 0.734 \\
        \midrule
        \multirow{2}{*}{Up Proj.}     & First  & 0.978  & 0.064  & 0.605 & 0.766 \\
                                      & Last   & 0.966  & 0.198  & 0.666 & 0.684 \\
        \midrule
        \multirow{2}{*}{Gate Proj.}   & First  & 0.986  & 0.048  & 0.443 & 0.824 \\
                                      & Last   & 0.951  & 0.165  & 0.584 & 0.626 \\
        \midrule
        \multirow{2}{*}{Down Proj.}   & First  & 1.000  & 0.013  & 0.071 & 0.976 \\
                                      & Last   & 0.996  & 0.049  & 0.181 & 0.869 \\
        \bottomrule
    \end{tabular}
    } 
    \vspace{-0.5em}
    \caption{Effect of $\mathbf{LR}$ initialization strategies in CALDERA. Results are shown for Layer 1’s weight matrices from Llama2-7B over 15 iterations. We compare $\|\bQ\bX\|$ and $\|\bL\bR\bX\|$, both normalized by $\|\bW\bX\|$ (i.e., $\|\bQ\bX\|/\|\bW\bX\|$) at first and last iterations. L2-norm is denoted by $\|\cdot\|$.}
    \label{tab:lrinit_order_norm_layer1_app}
\end{table}

\renewcommand{\arraystretch}{1.2}  
\setlength{\tabcolsep}{5pt}  
\begin{table}[!ht]
    \centering
    \Large
    \resizebox{\columnwidth}{!}{  
    \begin{tabular}{c | c | cc | cc}
        \toprule
        \multirow{3}{*}[-0.5em]{\centering Weight Type} & 
        \multirow{3}{*}[-0.5em]{\centering Iteration} & 
        \multicolumn{4}{c}{Initialization Method} \\
        \cmidrule(lr){3-6}
         & & \multicolumn{2}{c|}{\textbf{0}}  
           & \multicolumn{2}{c}{LRApprox($\mathbf{W}$)} \\  
        \cmidrule(lr){3-4} \cmidrule(lr){5-6}
         & & $\|\bQ\bX\|$ & $\|\bL\bR\bX\|$
           & $\|\bQ\bX\|$ & $\|\bL\bR\bX\|$ \\
        \midrule
        \multirow{2}{*}{Key Proj.}    & First  & 0.995  & 0.040  & 0.278  & 0.869 \\
                                      & Last   & 0.960  & 0.106  & 0.552  & 0.575 \\
        \midrule
        \multirow{2}{*}{Query Proj.}  & First  & 0.992  & 0.049  & 0.324  & 0.847 \\
                                      & Last   & 0.960  & 0.127  & 0.549  & 0.606 \\
        \midrule
        \multirow{2}{*}{Value Proj.}  & First  & 0.976  & 0.094  & 0.570  & 0.792 \\
                                      & Last   & 0.966  & 0.220  & 0.652  & 0.722 \\
        \midrule
        \multirow{2}{*}{O Proj.}      & First  & 0.978  & 0.102  & 0.644  & 0.733 \\
                                      & Last   & 0.970  & 0.265  & 0.725  & 0.680 \\
        \midrule
        \multirow{2}{*}{Up Proj.}     & First  & 0.978  & 0.077  & 0.621  & 0.740 \\
                                      & Last   & 0.978  & 0.182  & 0.676  & 0.686 \\
        \midrule
        \multirow{2}{*}{Gate Proj.}   & First  & 0.987  & 0.057  & 0.475  & 0.786 \\
                                      & Last   & 0.975  & 0.139  & 0.606  & 0.608 \\
        \midrule
        \multirow{2}{*}{Down Proj.}   & First  & 0.971  & 0.090  & 0.769  & 0.591 \\
                                      & Last   & 0.971  & 0.208  & 0.814  & 0.552 \\
        \bottomrule
    \end{tabular}
    } 
    \vspace{-0.5em}
    \caption{Effect of $\mathbf{LR}$ initialization strategies in CALDERA. Results are shown for Layer 10's weight matrices from Llama2-7B over 15 iterations. We compare $\|\bQ\bX\|$ and $\|\bL\bR\bX\|$, both normalized by $\|\bW\bX\|$ (i.e., $\|\bQ\bX\|/\|\bW\bX\|$) at first and last iterations. L2-norm is denoted by $\|\cdot\|$.}
    \label{tab:lrinit_order_norm_layer10_app}
\end{table}

\Cref{tab:lrinit_order_norm_layer1_app} and~\Cref{tab:lrinit_order_norm_layer10_app} present how different $\bL\bR$ initialization strategies affect the final weight distributions in Layer 1 and Layer 10, respectively. In both layers, we observe that varying the initialization leads to significantly different outcomes across all weights.
This finding reveals that joint optimization outcomes are highly sensitive to initialization choices, which ultimately determine whether quantization or matrix factorization dominates the final representation.

\subsection{Effect of LR Initialization on Joint Optimization for Q+LR} \label{app: scale and error}
\begin{figure*}[!htb]
    \centering
    \includegraphics[width=0.7\textwidth]{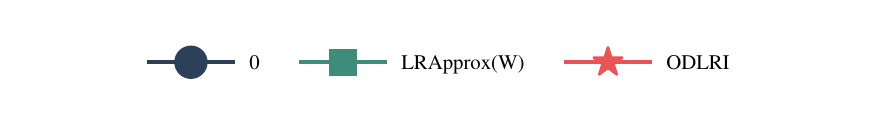}
    \begin{subfigure}{0.32\textwidth}
        \centering
        \includegraphics[width=\linewidth]{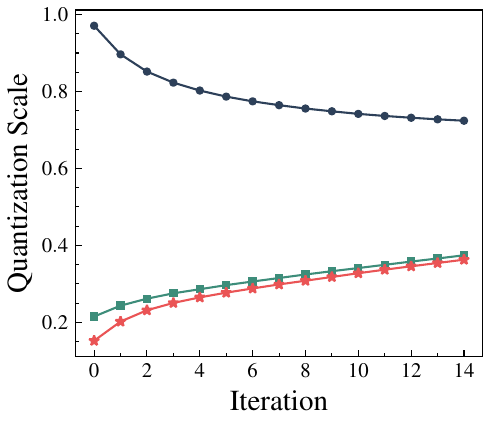}
        \caption{Layer 0: Key projection}
        \label{fig:q_scales_0_Key}
    \end{subfigure}
    \begin{subfigure}{0.32\textwidth}
        \centering
        \includegraphics[width=\linewidth]{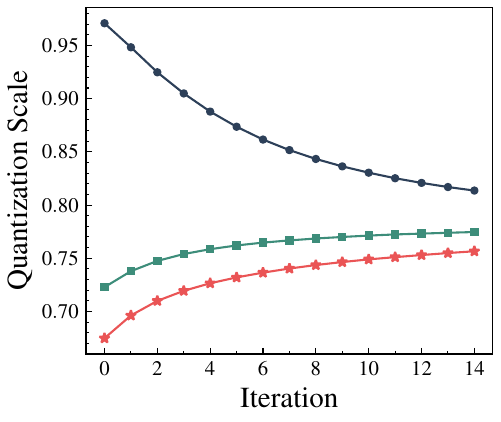}
        \caption{Layer 0: Value projection}
        \label{fig:q_scales_0_Value}
    \end{subfigure}
    \begin{subfigure}{0.32\textwidth}
        \centering
        \includegraphics[width=\linewidth]{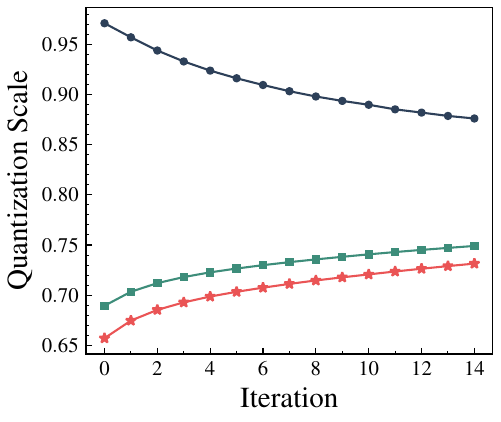}
        \caption{Layer 0: O Projection}
        \label{fig:q_scales_0_O}
    \end{subfigure}
    
    \begin{subfigure}{0.32\textwidth}
        \centering
        \includegraphics[width=\linewidth]{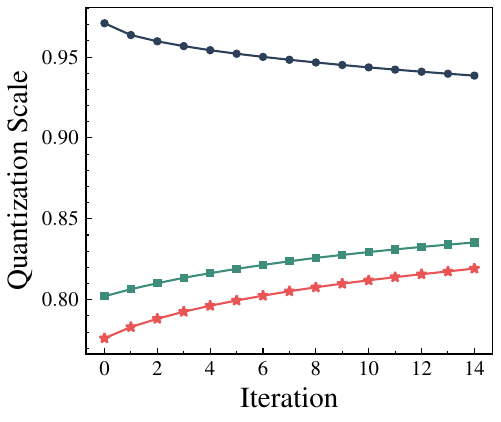}
        \caption{Layer 0: Gate projection}
        \label{fig:q_scales_0_Gate}
    \end{subfigure}
    \begin{subfigure}{0.32\textwidth}
        \centering
        \includegraphics[width=\linewidth]{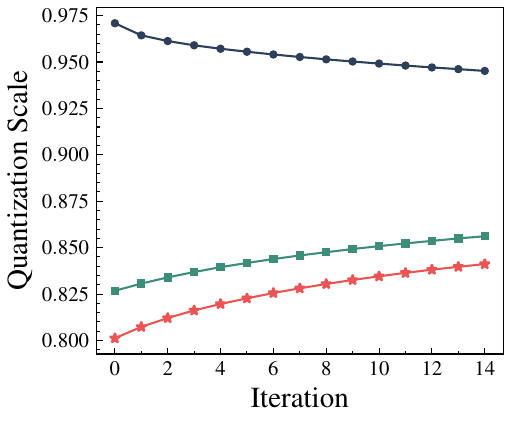}
        \caption{Layer 0: Up projection}
        \label{fig:q_scales_0_Up}
    \end{subfigure}
    \begin{subfigure}{0.32\textwidth}
        \centering
        \includegraphics[width=\linewidth]{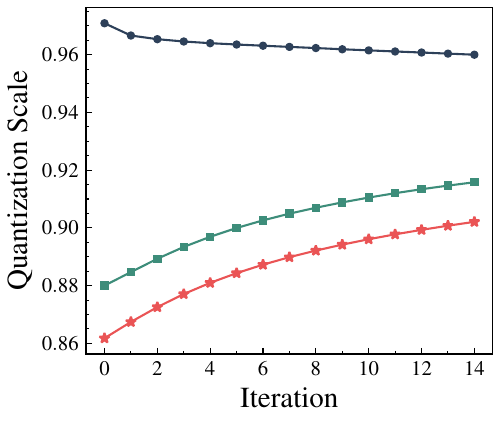}
        \caption{Layer 0: Down Projection}
        \label{fig:q_scales_0_Down}
    \end{subfigure}
    \begin{subfigure}{0.32\textwidth}
        \centering
        \includegraphics[width=\linewidth]{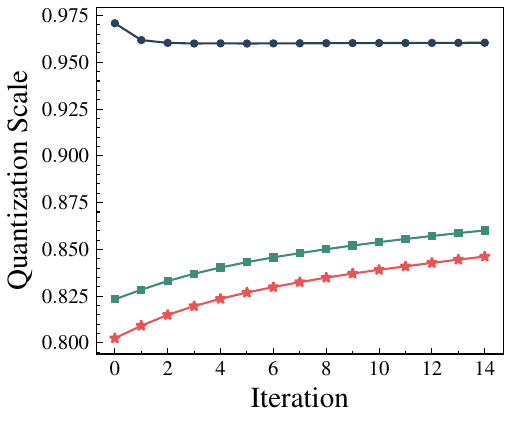}
        \caption{Layer 30: Key projection}
        \label{fig:q_scales_30_Key}
    \end{subfigure}
    \begin{subfigure}{0.32\textwidth}
        \centering
        \includegraphics[width=\linewidth]{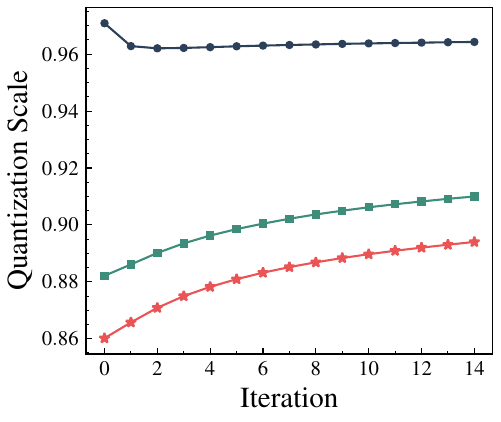}
        \caption{Layer 30: Value projection}
        \label{fig:q_scales_30_Value}
    \end{subfigure}
    \begin{subfigure}{0.32\textwidth}
        \centering
        \includegraphics[width=\linewidth]{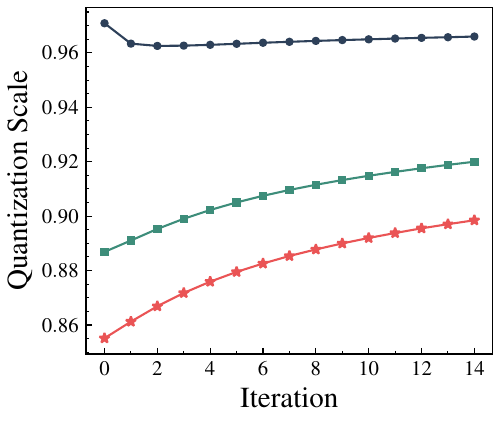}
        \caption{Layer 30: O projection}
        \label{fig:q_scales_30_O}
    \end{subfigure}
    
    \begin{subfigure}{0.32\textwidth}
        \centering
        \includegraphics[width=\linewidth]{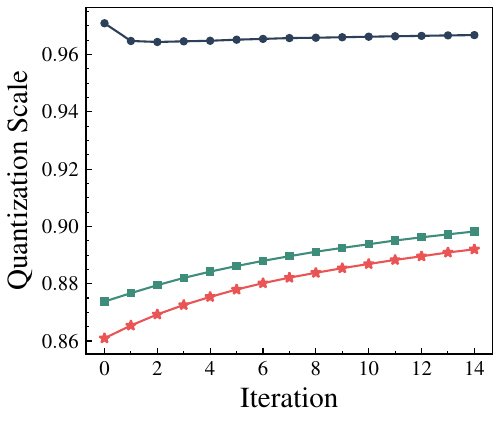}
        \caption{Layer 30: Gate projection}
        \label{fig:q_scales_30_Gate}
    \end{subfigure}
    \begin{subfigure}{0.32\textwidth}
        \centering
        \includegraphics[width=\linewidth]{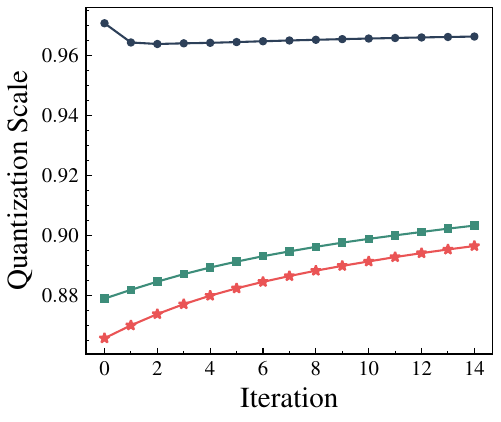}
        \caption{Layer 30: Up projection}
        \label{fig:q_scales_30_Up}
    \end{subfigure}
    \begin{subfigure}{0.32\textwidth}
        \centering
        \includegraphics[width=\linewidth]{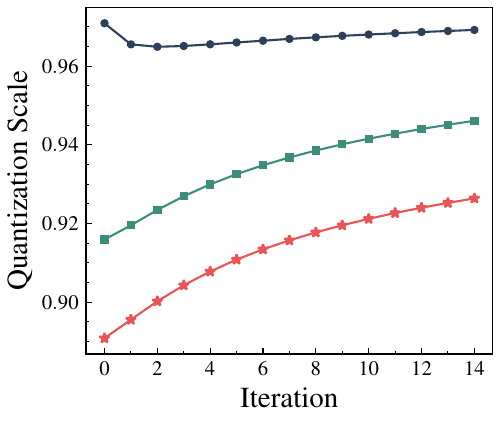}
        \caption{Layer 30: Down projection}
        \label{fig:q_scales_30_Down}
    \end{subfigure}

    \caption{\textbf{Quantization Scale across different initialization strategies.}
   We present the quantization scale over 15 iterations, where both $\bL$ and $\bR$ are quantized to 4-Bit at rank 256.
    Subplots display results for the Key, Value, O, Gate, Up, and Down projection layers in Layer 0 and Layer 30 of Llama2-7B.
    ODLRI (red stars) consistently achieves the lowest quantization scale, highlighting its effectiveness in low-bit quantization.}
    \vspace{20em}
    \label{fig:qscale_comp_app}
\end{figure*}

\Cref{fig:qscale_comp_app} examines the quantization scale across various layers, demonstrating that our method, ODLRI, effectively maintains optimal scale control throughout the network. The results indicate that ODLRI consistently outperforms baseline approaches in ensuring robust and stable quantization across diverse layers.

\begin{figure*}[!htb]
    \centering
    \includegraphics[width=0.7\textwidth]{figure/raw/legend}
    \begin{subfigure}{0.32\textwidth}
        \centering
        \includegraphics[width=\linewidth]{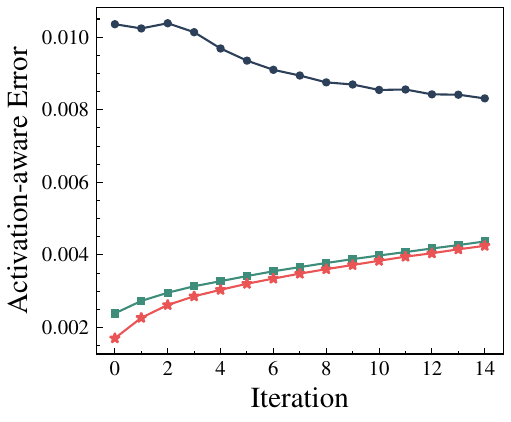}
        \caption{Layer 0: Key projection}
        \label{fig:min_error_0_Key}
    \end{subfigure}
    \begin{subfigure}{0.32\textwidth}
        \centering
        \includegraphics[width=\linewidth]{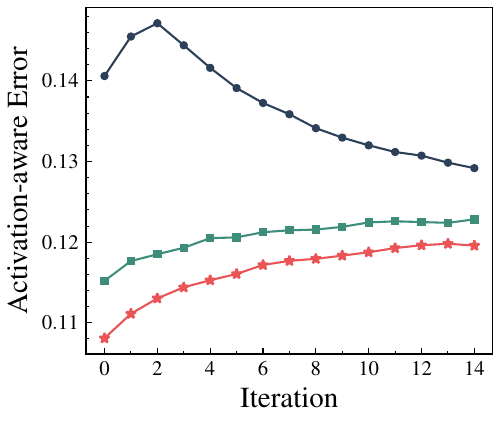}
        \caption{Layer 0: Value projection}
        \label{min_error_0_Value}
    \end{subfigure}
    \begin{subfigure}{0.32\textwidth}
        \centering
        \includegraphics[width=\linewidth]{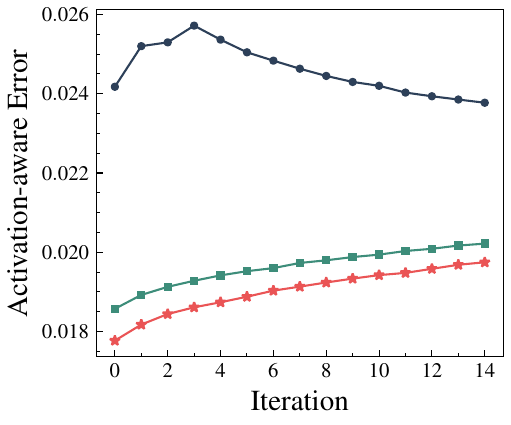}
        \caption{Layer 0: O Projection}
        \label{fig:min_error_0_O}
    \end{subfigure}
    
    \begin{subfigure}{0.32\textwidth}
        \centering
        \includegraphics[width=\linewidth]{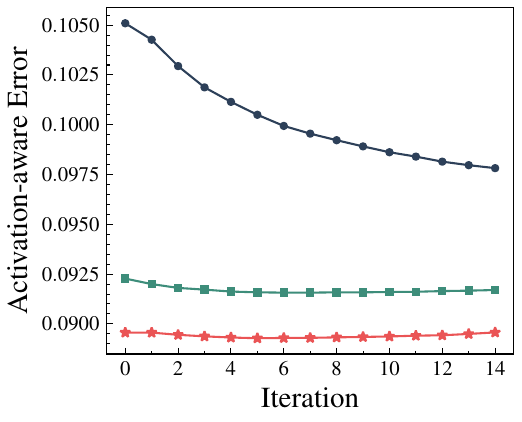}
        \caption{Layer 0: Gate projection}
        \label{fig:min_error_0_Gate}
    \end{subfigure}
    \begin{subfigure}{0.32\textwidth}
        \centering
        \includegraphics[width=\linewidth]{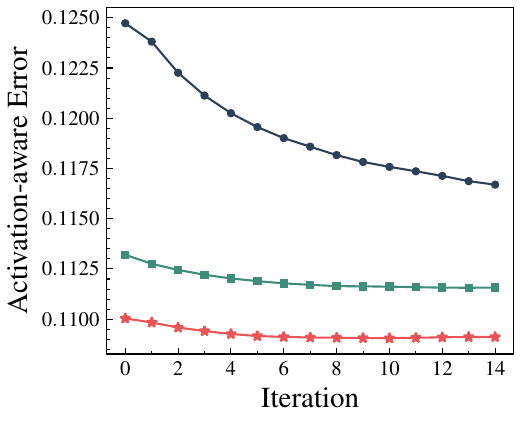}
        \caption{Layer 0: Up projection}
        \label{fig:min_error_0_Up}
    \end{subfigure}
    \begin{subfigure}{0.32\textwidth}
        \centering
        \includegraphics[width=\linewidth]{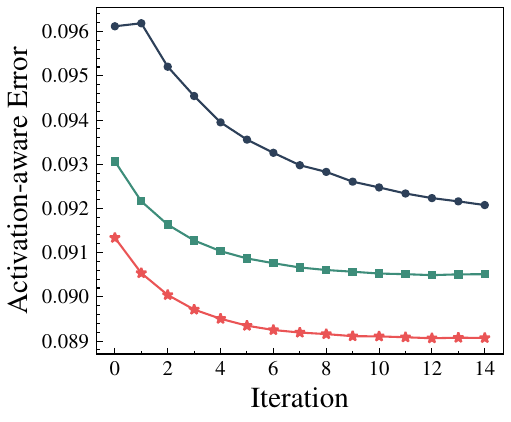}
        \caption{Layer 0: Down Projection}
        \label{fig:min_error_0_Down}
    \end{subfigure}
    \begin{subfigure}{0.32\textwidth}
        \centering
        \includegraphics[width=\linewidth]{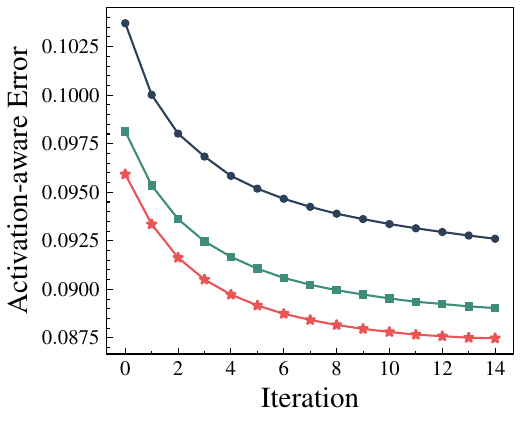}
        \caption{Layer 30: Key projection}
        \label{fig:min_error_30_Key}
    \end{subfigure}
    \begin{subfigure}{0.32\textwidth}
        \centering
        \includegraphics[width=\linewidth]{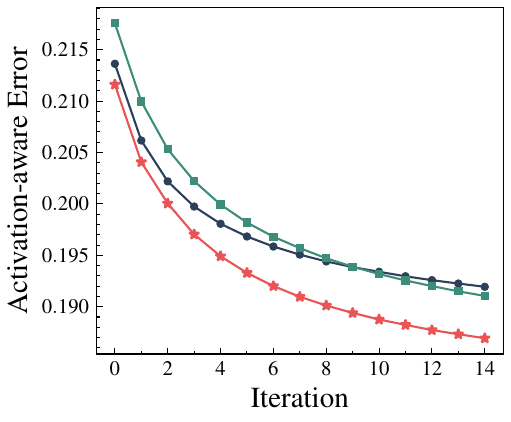}
        \caption{Layer 30: Value projection}
        \label{fig:min_error_30_Value}
    \end{subfigure}
    \begin{subfigure}{0.32\textwidth}
        \centering
        \includegraphics[width=\linewidth]{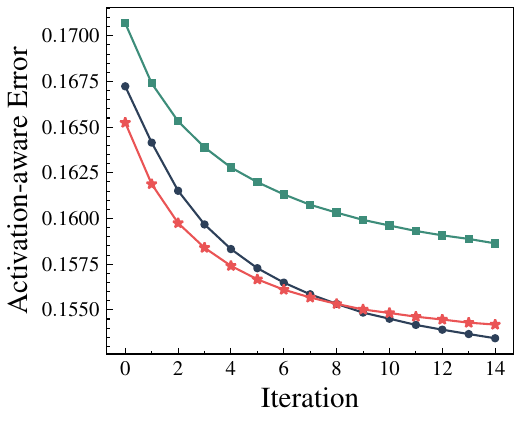}
        \caption{Layer 30: O projection}
        \label{fig:min_error_30_O}
    \end{subfigure}
    
    \begin{subfigure}{0.32\textwidth}
        \centering
        \includegraphics[width=\linewidth]{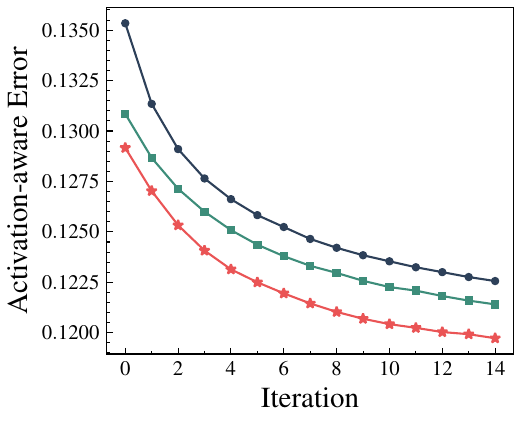}
        \caption{Layer 30: Gate projection}
        \label{fig:min_error_30_Gate}
    \end{subfigure}
    \begin{subfigure}{0.32\textwidth}
        \centering
        \includegraphics[width=\linewidth]{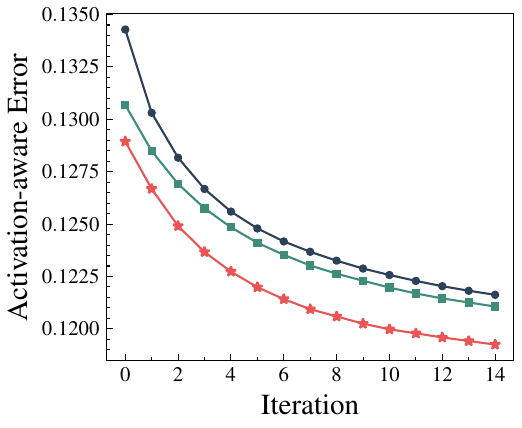}
        \caption{Layer 30: Up projection}
        \label{fig:min_error_30_Up}
    \end{subfigure}
    \begin{subfigure}{0.32\textwidth}
        \centering
        \includegraphics[width=\linewidth]{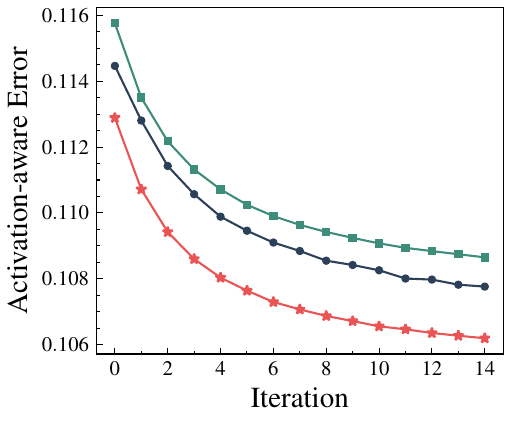}
        \caption{Layer 30: Down projection}
        \label{fig:min_error_30_Down}
    \end{subfigure}
    
    \caption{\textbf{Activation-aware Error across different initialization strategies.}~We present the normalized activation-aware error ${\|(\bW - \bQ - \bL \bR)\bX\|^2_\text{F}} / {\|\bW\bX\|^2_\text{F}}$
    over 15 iterations, where both $\bL$ and $\bR$ are quantized to 4-Bit at rank 256.
    Lower values indicate better preservation of activation information after quantization.
    Subplots display results for the Key, Value, O, Gate, Up, and Down projection layers in Layer 0 and Layer 30 of Llama2-7B.
    ODLRI (red stars) consistently shows the lowest error, demonstrating its effectiveness in reducing quantization error.}
    \vspace{20em}
    \label{fig:error_comp_app_l0}
\end{figure*}

Similarly,~\Cref{fig:error_comp_app_l0} presents the activation-aware error measured across multiple layers, confirming that ODLRI reliably minimizes activation-aware error in the preservation of intermediate activations. These findings underscore the efficacy of our approach in reducing activation-aware error, thereby enhancing overall model performance compared to baseline methods.


\end{document}